\pdfoutput=1

\documentclass[11pt]{article}

\usepackage[preprint]{acl}

\usepackage{times}
\usepackage{latexsym}
\usepackage{amsfonts}
\usepackage{subfigure}
\usepackage{subcaption}
\usepackage{graphicx}
\usepackage{amsmath} 
\usepackage{multirow}
\usepackage{xcolor}
\usepackage{multirow}
\usepackage{colortbl}
\usepackage{booktabs}
\usepackage{lipsum}
\usepackage[T1]{fontenc}

\usepackage[utf8]{inputenc}

\usepackage{microtype}

\usepackage{inconsolata}

\usepackage{graphicx}
\usepackage{tabularray}
\usepackage{soul}

\title{CARMA: Enhanced \textbf{C}ompositionality in LLMs via \textbf{A}dvanced \textbf{R}egularisation and \textbf{M}utual Information \textbf{A}lignment}

\author{Nura Aljaafari$^{1,4\dagger}$,~ Danilo S. Carvalho$^{1,3}$,~ Andr\'{e} Freitas$^{1,2,3}$ \\
  $^{1}$ Department of Computer Science, University of Manchester, United Kingdom\\
  $^{2}$ Idiap Research Institute, Switzerland\\
  $^{3}$ National Biomarker Centre, CRUK-MI, Univ. of Manchester, United Kingdom\\
  \texttt{\{firstname.lastname\}@[postgrad.]$^{\dagger}$manchester.ac.uk}}

\begin{document}
\maketitle

\begin{abstract}
Large language models (LLMs) struggle with compositional generalisation, limiting their ability to systematically combine learned components to interpret novel inputs. 
While architectural modifications, fine-tuning, and data augmentation improve compositionality, they often have limited adaptability, face scalability constraints, or yield diminishing returns on real data.
To address this, we propose CARMA, an intervention that enhances the stability and robustness of compositional reasoning in LLMs while preserving fine-tuned performance. CARMA employs mutual information regularisation and layer-wise stability constraints to mitigate feature fragmentation, ensuring structured representations persist across and within layers. We evaluate CARMA on inverse dictionary modelling and sentiment classification, measuring its impact on semantic consistency, performance stability, and robustness to lexical perturbations.
Results show that CARMA reduces the variability introduced by fine-tuning, stabilises token representations, and improves compositional reasoning. While its effectiveness varies across architectures, CARMA's key strength lies in reinforcing learned structures rather than introducing new capabilities, making it a scalable auxiliary method. These findings suggest that integrating CARMA with fine-tuning can improve compositional generalisation while maintaining task-specific performance in LLMs.
\end{abstract}

\section{Introduction}\label{sec:intro}
Compositional generalisation (CG) refers to the ability to systematically combine known expressions to generate novel ones following learned rules \cite{Partee1984}. This capability is essential for advancing language models (LMs) towards robust linguistic understanding beyond mere pattern matching \cite{ram2024makes}. 

Despite their strong performance across various NLP tasks, large language models (LLMs) exhibit persistent weaknesses in compositional generalisation \cite{ijcai2020p708, kim2020cogs, aljaafari2024interpreting}. These limitations stem from multiple factors, including training objectives and model architectures. Standard autoregressive training methods, such as next-token prediction, prioritise statistical correlations in token sequences over structured semantic understanding \cite{yin2023consistency, dziri2024faith}. As a result, token representations often lack structured compositionality, leading to fragmented information processing within layers (horizontal misalignment) and across layers (vertical inconsistency). 

Additionally, while self-attention mechanisms in Transformer models effectively capture local dependencies, they frequently fail to maintain coherent compositional representations across multiple layers \cite{murty-etal-2023-pushdown}. This misalignment impairs the model's ability to generalise compositionally, resulting in sensitivity to input order \cite{ismayilzada2024evaluating} and difficulties in handling complex syntactic and morphological structures \cite{aljaafari2024interpreting}. 

\begin{figure*}
    \centering
    \includegraphics[width=.95\linewidth]{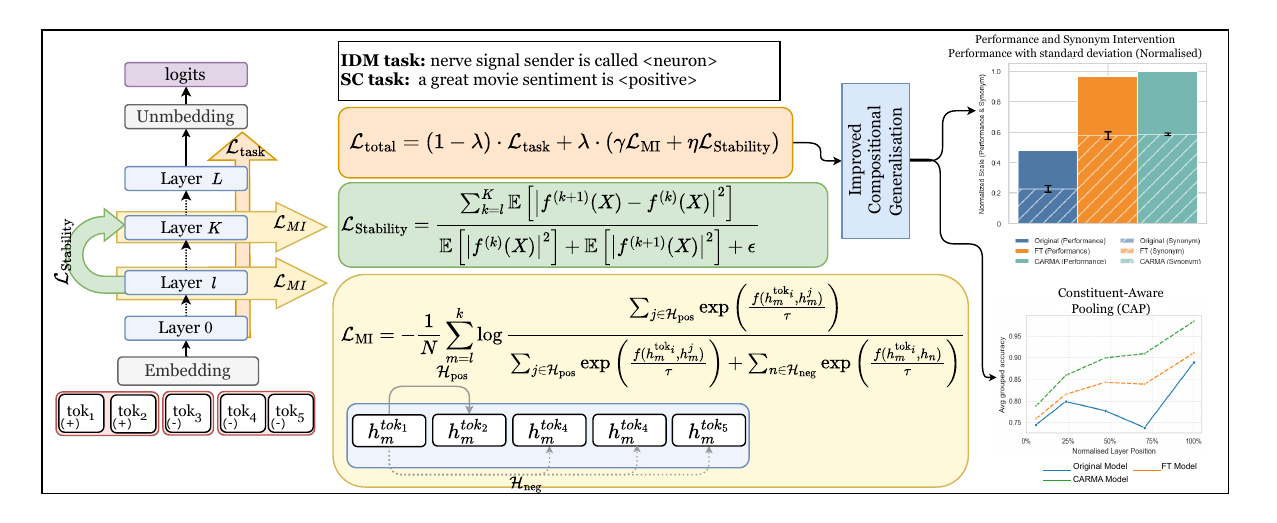}
 \caption{
This diagram depicts the computation of the loss and illustrates the integration of the \textbf{Mutual Information (MI) loss} (\(\mathcal{L}_{\text{MI}}\)) and the \textbf{Stability Loss} (\(\mathcal{L}_{\text{stability}}\)) into the final optimisation process. Tokens \(Tok_1\) and \(Tok_2\) form the \textit{positive set} (\(H_{\text{pos}}\)), while \(Tok_3, Tok_4, Tok_5\) form the \textit{negative set} (\(H_{\text{neg}}\)). The \(\mathcal{L}_{\text{MI}}\) loss is computed \textit{vertically} across layers (\(l\) to \(k\)), maximising the similarity of tokens in \(H_{\text{pos}}\) while contrasting them with tokens in \(H_{\text{neg}}\). The \(\mathcal{L}_{\text{stability}}\) loss is computed \textit{horizontally} between consecutive layers, ensuring consistency in hidden state representations. Both auxiliary losses are combined with the task loss (\(\mathcal{L}_{\text{task}}\)) to form the total loss (\(\mathcal{L}_{\text{total}}\)). This integration improves token representations and enhances the model's overall optimisation.}
    \label{fig:CCG}
\end{figure*}

Several approaches have been proposed to address these limitations, including architectural modifications, enhanced encoding strategies, and targeted regularisation techniques \cite{ontanon-etal-2022-making, murty-etal-2023-pushdown, csordas-etal-2021-devil}. However, these methods often struggle to balance compositional improvements with maintaining performance across diverse downstream tasks. Moreover, their effectiveness is typically confined to specific compositional structures or synthetic benchmarks. Developing a robust and adaptable solution that enables LLMs to achieve consistent CG across diverse tasks remains a major challenge.

This work introduces \textbf{CARMA}: enhanced \textbf{C}ompositionality in LLMs via \textbf{A}dvanced \textbf{R}egularisation and \textbf{M}utual Information \textbf{A}lignment,  illustrated in Figure~\ref{fig:CCG}. CARMA enhances CG by addressing training challenges that hinder structured compositionality in LLMs. By balancing layer-specific updates and reinforcing token-level dependencies, CARMA provides a scalable and adaptable solution that improves CG without sacrificing downstream task performance. To evaluate CARMA's effectiveness, we investigate the following research questions:  

\begin{itemize}
    \item \textbf{RQ1:} How does regulating mutual information across layers influence compositionality in LLMs? How does it affect sensitivity to input and internal perturbations?  
    \item \textbf{RQ2:} To what extent does layer-specific regularisation improve compositional generalisation across semantic and sentiment analysis tasks, assessing CARMA's adaptability across domains?
\end{itemize}

\noindent The key contributions of this work are as follows: 

\begin{itemize}
    \item A novel regularisation method that enhances compositional generalisation without requiring architectural modifications. CARMA leverages mutual information alignment to preserve token dependencies across layers and employs layer-wise stability constraints to reduce representational inconsistencies. 
    
    \item A systematic evaluation of CARMA across compositionally demanding tasks, demonstrating its ability to reinforce systematicity and substitutivity, particularly in models where fine-tuning alone is insufficient.  
    \item A theoretical and empirical analysis of how token dependencies degrade across layers in standard LLMs, revealing that CG limitations are not solely dependent on model size but rather on representational instability. CARMA mitigates this by ensuring consistent information flow, showing that non-intrusive regularisation strategies can significantly improve CG. 
\end{itemize}
The remainder of this paper is structured as follows: Section~\ref{sec:compositionality} reviews compositionality in LLMs and associated challenges. Section~\ref{sec:methodology} introduces the CARMA method. Section~\ref{sec:experiments} describes the experimental setup. Section~\ref{sec:results} presents empirical findings. Section~\ref{sec:related} discusses related work. Section~\ref{sec:conclusion} offers insights and future research directions. Supporting datasets and software are available at a public repository.\footnote{Anonymised for review.}

\section{Compositionality in LLMs}\label{sec:compositionality}
Compositional generalisation (CG) in linguistics encompasses five key principles: systematicity, productivity, substitutivity, localism, and over-generalisation \cite{dankers-etal-2022-paradox}. These principles have been explored in LLMs across compositional instruction \cite{yang2024exploring}, semantic parsing \cite{li-etal-2023-learning}, translation \cite{li-etal-2021-compositional}, and multi-step inference \cite{zhang2024can}. Studies show standard Transformer-based LLMs exhibit limited CG, struggling with basic compositional tasks such as assembling tokens into words or constructing morphemes \cite{aljaafari2024interpreting, ismayilzada2024evaluating}. These limitations are linked to architectural constraints, training objectives, and tokenisation practices that fragment information and increase sensitivity to input order and contextual noise \cite{murty-etal-2023-pushdown}.

\textbf{Training Objectives and Information Fragmentation.} Standard training objectives for LLMs typically optimise for next-token prediction, which prioritises surface-level correlations over deeper semantic integration \cite{dziri2024faith}. While this approach is effective for data already seen, it often impedes CG by reducing mutual information between dependent tokens, thereby limiting the model's ability to form coherent compositional representations \cite{aljaafari2024interpreting}.

\textbf{Architectural Mechanisms and Compositional Consistency.} Beyond training objectives, architectural mechanisms such as dropout and self-attention disperse information across the model, increasing sensitivity to input order and context. This undermines \textbf{compositional consistency} \cite{sajjadi2016regularization, cai2021isotropy}, the ability to maintain consistent outputs when processing semantically equivalent inputs through transformations like word substitution or paraphrasing. These challenges impact both high-complexity reasoning tasks and simpler operations that demand consistent morphological and syntactic processing \cite{ismayilzada2024evaluating}.

\textbf{Existing Approaches to Enhance CG in LLMs.} Research has explored architectural adjustments, regularisation techniques, and task-specific strategies to address CG limitations. \citet{ontanon-etal-2022-making} demonstrated that combining relative positional encoding with embeddings enhances CG in algorithmic tasks, while weight sharing and copy decoders help retain input structures. Architectural modifications like Pushdown Layers \cite{murty-etal-2023-pushdown} and GroCoT \cite{sikarwar-etal-2022-transformers} incorporate mechanisms for tracking syntactic depth and spatial relations, enabling recursive processing of compositional structures. RegularGPT \cite{chi-etal-2023-transformer} introduces adaptive depth and memory mechanisms to facilitate CG. Studies by \citet{csordas-etal-2021-devil} and \citet{petty2024impact} reveal that architectural choices and training setups significantly impact CG enhancement. In neural machine translation, \citet{dankers2022paradox} found a positive correlation between data size and compositional performance.\\
Frameworks like CompMCTG and Meta-MCTG \cite{zhong2024benchmarking} suggest joint training and meta-learning approaches improve fluency, though performance drops persist in out-of-distribution tasks. Synthetic tasks show recursive, step-by-step prompt formats support combinatorial generalisation, despite training biases and sequence order constraints \cite{ramesh2024compositional}.

\section{Enhanced Compositionality via Advanced Regularisation and Mutual Information Alignment (CARMA)}\label{sec:methodology}
This section formalises compositionality, introduces the core principles of CARMA, and details its components. Figure \ref{fig:CCG} illustrates the method, highlighting its process and key components.

\subsection{Compositionality Formalisation}
\paragraph{Mathematical Foundations of Compositionality.} CG (Section~\ref{sec:compositionality}) can be formally defined through a compositional system where $\mathcal{E}$ denotes a set of expressions (e.g., token sequences recognised by the model), and $\mathcal{M}$ represents a corresponding set of meanings. This relationship is formalised as a function:
\begin{equation}\label{eq:cg_system_1}
f: \mathcal{E} \rightarrow \mathcal{M}
\end{equation}
For any complex expression \( e \in \mathcal{E} \), composed of constituent elements \( e_1, \dots, e_n \) according to a syntactic rule \( r \), the function \( f \) satisfies:
\begin{equation}\label{eq:cg_system_2}
f(r(e_1, \dots, e_n)) = g_r(f(e_1), \dots, f(e_n)),
\end{equation}
where $g_r$ is the semantic operation that corresponds to the syntactic rule $r$.
\paragraph{Compositional Generalisation in LLMs.} Effective CG in LLMs requires generating structured compositions that preserve semantic consistency. Given a novel expression $e_{\text{novel}}$ similar to a known expression $e_{\text{known}}$ within a threshold $\beta$, their semantic functions must remain within an interpretable bound or deviation $\alpha$:
\begin{equation}
d(e_{\text{novel}}, e_{\text{known}}) \leq \beta \Rightarrow d(f(e_{\text{novel}}), f(e_{\text{known}})) \leq \alpha.
\end{equation}
This formulation captures systematicity (structured combinations), substitutivity (preservation under transformations), and resistance to over-generalisation (bounded semantic deviation) while maintaining interpretability.

\subsection{CARMA Formalisation}
CARMA operates over a range of target layers, from \( l \) to \( \mathcal{K}\) (\( 0 < l \leq \mathcal{K} \leq L \), where \( L \) is the total number of layers), and consists of two core components: \textbf{Mutual Information} and \textbf{Layer-Wise Stability} Regularisation.

\paragraph{Mutual Information (MI) Regularisation Across Layers.} CARMA preserves essential dependencies and maintains structural coherence by maximising MI between hidden states of related tokens. The MI between hidden states \( h^k_i \) and \( h^k_j \) at layer \( k \), representing two related tokens \( i \) and \( j \), is defined as:
\begin{equation} 
I(h^k_i; h^k_j) = \mathbb{E}_{P(h^k_i,h^k_j)}\left[\log \frac{P(h^k_i,h^k_j)}{P(h^k_i)P(h^k_j)}\right]
\end{equation}
Since exact computation is intractable, MI is approximated using the InfoNCE loss \cite{oord2018representation}, encouraging token-level dependencies across the same layers:
\begin{equation}
\small
\begin{aligned}
    \mathcal{L}_{\text{MI}} = - \frac{1}{N} \sum_{k=l}^{\mathcal{K}} \sum_{i=1}^Q \Bigg( 
    &\log \sum_{\substack{h_j \in \mathcal{H}^k \\ j \neq i}} \exp\left(\frac{f(h_i^k, h_j^k)}{\tau}\right) \\
    &- \log \Bigg( \sum_{\substack{h_j \in \mathcal{H}^k \\ j \neq i}} \exp\left(\frac{f(h_i^k, h_j^k)}{\tau}\right) \\
    &\quad + \sum_{h_m \in \mathcal{N}^k} \exp\left(\frac{f(h_i^k, h_m)}{\tau}\right) \Bigg) \Bigg),
\end{aligned}
\end{equation}
where \( f(h_i^k, h_j^k) \) is a similarity function quantifying the relationship between hidden states at layer \( k \), \( \mathcal{H}^k \) denotes the set of positive examples related to \( h_i^k \), \( \mathcal{N}^k \) is the set of negative examples unrelated to \( h_i^k \) at layer \( k \), \( \tau \) is the temperature parameter, and \( N \) is the total number of target layers from \( l \) to \( \mathcal{K} \), with \( Q \) representing the number of tokens or samples used per layer. Further details on MI approximation are provided in Appendix~\ref{sec:infoNCE_MI}.

\paragraph{Layer-Wise Stability Regularisation.}
This component enforces smooth transitions across layers, reducing abrupt changes that could disrupt compositional structures. For a layer \( k \), the Layer-Wise Stability Loss is defined as:
\begin{equation}
\small
\mathcal{L}_{\text{Stability}} = \sum_{k=l}^{\mathcal{K}} \mathbb{E} \left[ \frac{\left| f^{(k+1)}(X) - f^{(k)}(X) \right|^2}{\mathbb{E}\left[ \left| f^{(k)}(X) \right|^2 \right] + \mathbb{E}\left[ \left| f^{(k+1)}(X) \right|^2 \right] + \epsilon} \right],
\end{equation}
where \( f^{(k)}(X) \) denotes the output of layer \( k \) (i.e., after the attention and MLP submodules), and \( \epsilon \) is a small positive constant to ensure numerical stability (e.g., \( \epsilon = 10^{-8} \)). Minimising this loss preserves compositional integrity across the specified layers by encouraging smooth and consistent transitions between them, thereby enabling more stable information flow and aggregation within this range.

\paragraph{CARMA Loss.} CARMA integrates $\mathcal{L}_{\text{MI}}$ and $\mathcal{L}_{\text{Stability}}$ into its total loss as:
\begin{equation}\label{eq:carma_loss}
\mathcal{L}_{\text{CARMA}} = \gamma \mathcal{L}_{\text{MI}} + \eta \mathcal{L}_{\text{Stability}},
\end{equation}
where \( \gamma \) and \( \eta \) are hyperparameters in \( [0, 1] \) that control the relative contribution of each component. The final optimisation objective balances task-specific performance with CARMA’s regularisation as:
\begin{equation}
    \mathcal{L}_{\text{total}} = (1 - \lambda) \cdot \mathcal{L}_{\text{task}} + \lambda \cdot \mathcal{L}_{\text{CARMA}},
\end{equation}
where \( \mathcal{L}_{\text{task}} \) represents the task-specific loss, \( \mathcal{L}_{\text{CARMA}} \) is the regularisation loss, and \( \lambda \in [0, 1] \) controls the trade-off between task accuracy and compositional robustness.

\paragraph{Layer Selection for Regularisation.}
We apply \textsc{CARMA} to layers around one-third of the model depth, based on evidence that early-to-mid layers better capture compositional and syntactic structure, while deeper layers tend to specialise in task-specific representations~\citep{he-etal-2024-decoding,langedijk-etal-2024-decoderlens}. In our preliminary experiments, we observed that performance gains diminish when regularisation is applied to deeper layers. As a default, we recommend layers 3--4 in 12-layer models and 6--10 in 24-layer models.

\paragraph{Evaluation Metrics.} We use exact match accuracy as the primary metric for both IDM and SC. This choice is motivated by the fact that both tasks have closed and categorical output spaces: IDM outputs are limited to a predefined set of single-word lexical entries from WordNet, while SC uses a fixed sentiment label set. In such settings, exact match is a standard and appropriate evaluation criterion.

\section{Experimental Setup}\label{sec:experiments}
\subsection{Downstream Tasks \& datasets}
Two tasks that assess different aspects of compositional generalisation are used in the paper: \textbf{Inverse Dictionary Modelling (IDM)} for word-level composition and \textbf{Sentiment Classification (SC)} for phrase-level structure. These tasks measure systematicity, substitutivity, over-generalisation, and robustness to perturbations.\\
\noindent\textbf{IDM} evaluates a model's ability to generate terms from definitions, focusing on substitutivity in semantic composition. Using WordNet \cite{miller-1994-wordnet} with an 80-10-10 train-validation-test split, models are prompted with a definition to generate the corresponding term (e.g., The shore of a sea is called'' $\rightarrow$ coast'').  By mapping definitions to terms, this task provides a robust assessment of a model's ability to perform compositional substitution.\\
\noindent\textbf{SC} assesses the model's ability to infer sentiment from phrases and sentences, particularly focusing on sentiment shifts and over-generalisation. Using the Stanford Sentiment Treebank (SST) \cite{socher2013recursive} with its original splits, models predict sentiment labels from textual inputs (e.g., A brilliant performance sentiment is'' $\rightarrow$ positive''). This task examines how sentiment composition is preserved across different levels of linguistic structure.
For both tasks, performance is assessed using Exact Match Accuracy, providing a robust assessment of compositional substitution ability. Task formalisation, dataset details, and task selection rationale are in Appendices~\ref{sec:task_selection}, \ref{sec:task_formalisation}, and \ref{sec:dataset_appendix}, respectively.

\subsection{Models and Experimental setup}
We evaluate three setups: original models, task-specific fine-tuning, and fine-tuning with \textsc{CARMA} regularisation. We test GPT-2 (S/L) \cite{radford2019language}, Gemma1–2B \cite{team2024gemma}, LLaMA3.2 (1B/3B) \cite{dubey2024llama}, and Qwen2.5 (0.5B/3B) \cite{yang2024qwen2}. We focus on decoder-only architectures, as they represent the dominant paradigm in many open-weight and production-ready LLMs. \textsc{CARMA} is generally applied at approximately one-third of the model’s depth, though specific layer positions vary. Details on software, FT methodologies, model specifications, and \textsc{CARMA} hyperparameter selection are provided in Appendices~\ref{sec:fine_tuning_appenix} and \ref{sec:exp_setup}.


\subsection{Interventions for Compositional Robustness and Performance Stability}
Two interventions are used to evaluate the robustness of compositional structures and the stability of learned representations: Constituent-aware pooling (CAP) and synonym replacement. These interventions assess hierarchical dependencies and semantic consistency under controlled perturbations.

\noindent\textbf{CAP}~\cite{aljaafari2024interpreting} groups token-level representations into higher-level semantic units (e.g., words, constituents) to assess hierarchical dependencies and how compositional structures are maintained across layers. In this paper, the token-to-word CAP is utilised. Model robustness is measured by monitoring performance metrics before and after applying CAP. Full methodology and formalisation are provided in Appendix~\ref{sec:cap_explanation}.

\noindent\textbf{Synonym Replacement} evaluates semantic consistency by substituting 25\% and 40\% of prompt words with synonyms within an interpretable bound ($\alpha$). Experiments were repeated at least five times with different seeds for robustness and performance stability assessment; further details are in Appendix~\ref{sec:syn_replacement}.  

\begin{figure*}
    \centering
    \subfigure[IDM Task]{\label{fig:carma_results_idm}\includegraphics[width=.45\linewidth]{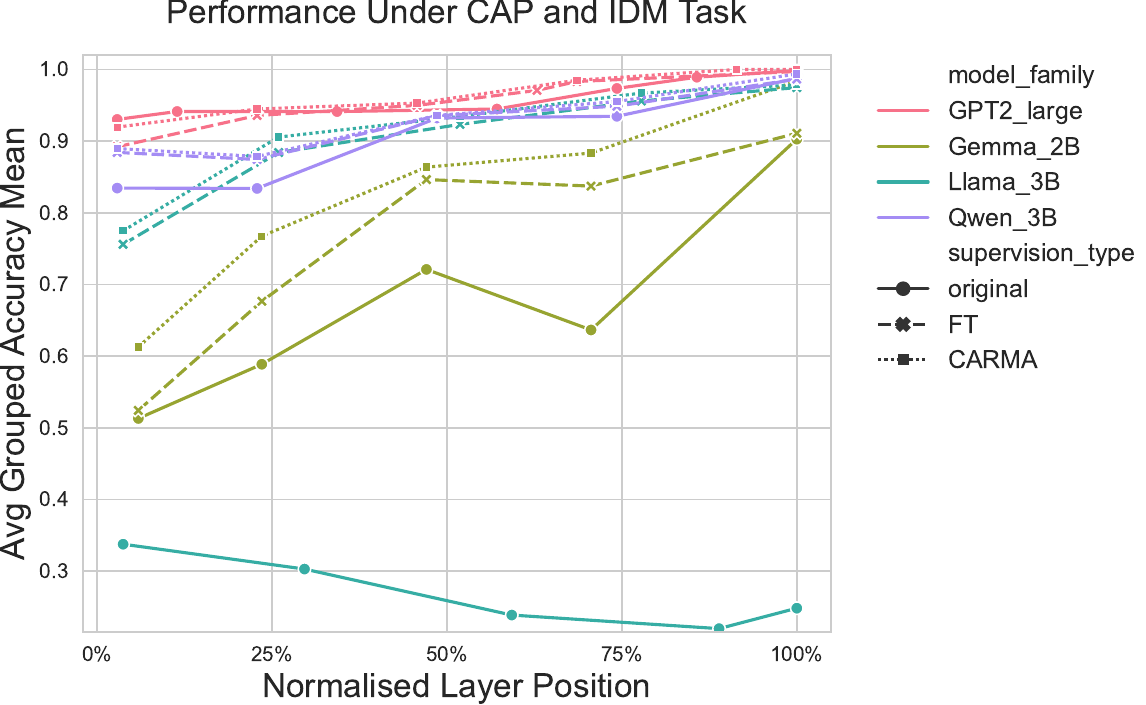}}
    \subfigure[SC Task]{\label{fig:carma_results_sst}\includegraphics[width=.45\linewidth]{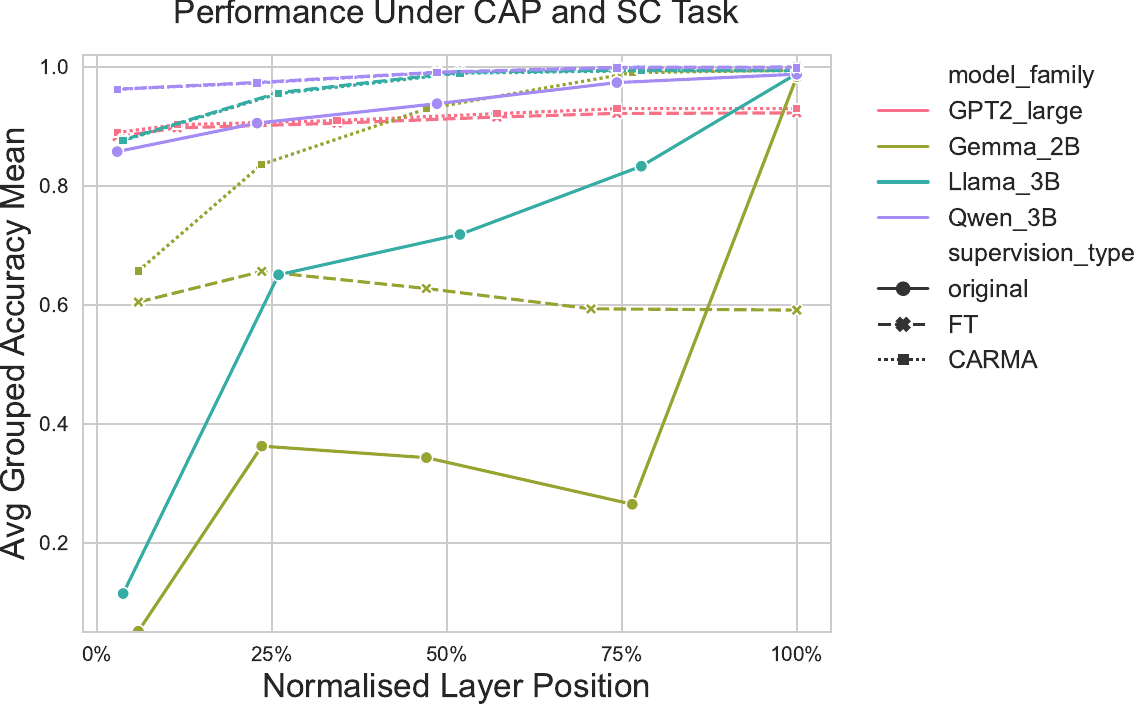}}
        \caption{Layer-wise performance comparison under CAP intervention, with performance averaged over three protocols (Mean CAP, Max CAP, Sum CAP) for Original, Fine-Tuned (FT), and CARMA (FT + CARMA) models. Layer numbers are normalised to their relative positions within each model to enable cross-architecture comparison. The IDM task (left) highlights CARMA's improvements in systematicity and stability, particularly in the early and middle layers. The SC task (right) demonstrates CARMA's ability to enhance robustness, though convergence with FT occurs in deeper layers.}

    \label{fig:carma_results_combined}
\end{figure*}

\section{Results and discussion}\label{sec:results}
The method is evaluated across three aspects its impact on: (1) model robustness against compositional-based perturbations, (2) model performance stability, and (3) model overall performance. See Appendix~\ref{sec:evaluation_metrics} for a detailed breakdown of the evaluation metrics used for each aspect.

\subsection{Constituent-Aware Pooling (CAP) Intervention}
Fig.~\ref{fig:carma_results_idm} and~\ref{fig:carma_results_sst} show the impact of CAP on both tasks, comparing original, fine-tuned (FT), and \textsc{CARMA} models.\footnote{Throughout this paper, models incorporating CARMA with FT are referred to as CARMA models.} Each plot shows performance across normalised layer positions, where Accuracy is averaged over three CAP protocols (Mean, Max, Sum); protocol-specific results and extended comparisons are included in Appendix~\ref{sec:additiona_results}. The analysis examines how well models preserve compositionality under hierarchical pooling.

CARMA's effectiveness is influenced by model size, tokenisation strategy, and task complexity. \ul{In IDM tasks, CARMA models have considerable gains when applying CAP at the earliest layers (1\% of model depth),} particularly in models with fine-grained tokenisation: Llama-1B (+3.61\%) and Gemma-2B (+16.89\%). GPT2-L, despite its reliance on subword tokenisation, benefits from CARMA over FT (+3.67\%). However, Llama-3B and Qwen-3B minimal improvements (+1.0\%) suggest a capacity ceiling where increased model size does not yield proportional gains due to training data limitations. \ul{The combination of smaller scale and multilingual training particularly affects Qwen-0.5B, where limited model capacity coupled with broad language coverage appears to constrain English-specific compositional learning,} resulting in reduced CARMA benefits. In SC tasks, tokenisation effects vary with task complexity. When intervening at 25\% layer position, Gemma-2B and Llama-1B show the strongest gains (+27.38\%, +10.59\%), while Llama-3B exhibits a marginal difference between CARMA and FT ($\sim1\%$) but still outperforms the Original model (+37.68\%). These results suggest that fine-tuning alone is sufficient for simpler tasks, whereas structured interventions like CARMA are particularly beneficial for more complex, compositional reasoning tasks.

\ul{In a layer-wise analysis, the impact of CARMA varies significantly across network depths, revealing crucial insights about compositional learning in transformers}. Early layers (0-25\%) benefit the most from regularisation, as they establish foundational compositional representations by exhibiting a weak notion of compositionality. Middle layers (25-75\%) reinforce these patterns, maintaining structured feature dependencies with moderate improvements. Deeper layers (75-100\%) show minimal benefits as the model transitions from compositional learning to task-specialised representations. This pattern aligns with previous findings on layer-wise compositional evolution in Transformers, where earlier layers capture hierarchical structure, while deeper layers exhibit increased task specificity \cite{feucht2024footprints}. CARMA can thus be strategically applied to control these early representations, maintaining beneficial compositional structure while allowing natural task-specific adaptations in deeper layers.\\
These findings demonstrate CARMA's effectiveness, particularly for models with granular tokenisation under data constraints, mediated by model capacity and task demands. The method's dual role - enhancing early compositional learning while preserving deeper layer adaptations - enables targeted improvement in model robustness without disrupting task-specific processing.

\begin{table}[h]
\centering
\tiny
\begin{tabular}{|l|c|c|c|c|c|}
\hline
\textbf{Model} & \textbf{Ver.} & \textbf{Task} & \textbf{Int.} & \textbf{CS} & \textbf{CV} \\
\hline
\multirow{5}{*}{GPT2-L} 
& CARMA & IDM & 25\% & 56.31 & \textbf{0.0164} \\
& FT & IDM & 25\% & \textbf{56.95} & 0.0311 \\
& Org & IDM & 25\% & 51.10 & 0.1175 \\
\cline{2-6}
& CARMA & SC & 25\% & \textbf{0.8858} & \textbf{0.0065} \\
& FT & SC & 25\% & 0.8804 & 0.0082 \\
\hline
\multirow{6}{*}{Gemma-2B} 
& CARMA & IDM & 25\% & {56.70} & \textbf{0.023} \\
& FT & IDM & 25\% & \textbf{57.42} & 0.030 \\
& Org & IDM & 25\% & 49.47 & 0.031 \\
\cline{2-6}
& CARMA & SC & 25\% & 78.90 & \textbf{0.008} \\
& FT & SC & 25\% & \textbf{80.23} & 0.009 \\
& Org & SC & 25\% & 68.14 & 0.042 \\
\hline
\multirow{6}{*}{Llama-3B} 
& CARMA & IDM & 25\% & \textbf{62.86} & \textbf{0.015} \\
& FT & IDM & 25\% & 62.22 & 0.029 \\
& Org & IDM & 25\% & 52.47 & 0.035 \\
\cline{2-6}
& CARMA & SC & 25\% & 84.83 & \textbf{0.0056} \\
& FT & SC & 25\% & \textbf{85.85} & 0.0065 \\
& Org & SC & 25\% & 35.21 &0.0136 \\
\hline
\end{tabular}
\caption{Model performance (25\% synonym intervention). \textbf{Ver.}: Version; \textbf{Int.}: Intervention rate; \textbf{CS}: ConsistSyn (\%); \textbf{CV}: Coefficient of Variation. \textbf{Best values in bold.}}
\label{tab:synonym_results}
\end{table}

\subsection{Synonyms Replacement Intervention} 
Synonym Replacement evaluates semantic consistency and robustness under lexical variations across multiple runs ($N\geq5$) with different seeds. \textit{ConsistSyn} measures output preservation after substitution, while the coefficient of variation (CV) quantifies performance stability, with lower values indicating higher stability. Performance is assessed at 25\% and 40\% word replacement rates to measure sensitivity to perturbations. Sample results are in Table~\ref{tab:synonym_results}; full details appear in Appendix~\ref{sec:additiona_results}.

Across models, CARMA achieves a distinctive performance profile, matching or exceeding FT \textit{ConsistSyn} while consistently demonstrating superior stability through lower CV values. At 25\% intervention, Gemma-2B CARMA achieves 56.70\% \textit{ConsistSyn} with a CV of 0.0225, compared to FT's 57.42\% with higher variance (CV: 0.0307). Llama-3B CARMA outperforms FT in both \textit{ConsistSyn} (62.86\% vs. 62.22\%) and stability (CV: 0.0148 vs. 0.0292) for IDM. Qwen-3B follows a similar trend but with smaller relative gains, improving stability (CV: 0.0225 vs. 0.0279) while maintaining a marginal \textit{ConsistSyn} advantage over FT (62.00\% vs. 61.79\%). However, as intervention complexity increases to 40\%, the performance gap widens; for example, Gemma-2B FT maintains higher \textit{ConsistSyn} (44.98\%) than CARMA (42.36\%), though CARMA remains more stable (CV: 0.0174 vs. 0.0249). This behaviour implies that the advantage of CARMA lies in its lower variance and reinforcement of compositional consistency. Thus, \ul{it maintains compositional understanding without sacrificing performance, whereas FT produces a performance-driven approach.}

While the absolute differences in \textit{ConsistSyn} between CARMA and FT are sometimes modest, particularly at lower replacement rates (e.g., 25\%), the stability benefits of CARMA become more evident under increased perturbation (e.g., 40\%), where FT models often show degraded consistency. \textit{In these higher-variance regimes, CARMA consistently reduces output variability across model families, reinforcing its utility as a robustness-oriented intervention, even when raw accuracy remains comparable.}

The tokenisation method significantly affects CARMA's impact. Models with more structured tokenisation show stronger stability improvements, but gains vary based on vocabulary design and language coverage. Llama and GPT2-L generally benefit more than Qwen, even with similar sizes, likely due to their smaller multilingual coverage, which results in a more compact and consistent token distribution. Qwen, with a larger vocabulary (151K tokens) supporting broader multilingual processing, introduces redundancy that dampens CARMA's relative stability advantage. Gemma-2B, optimised for a single dominant language with a large vocabulary size, shows the highest overall gains, reinforcing that a structured tokenisation approach focused on a limited linguistic scope enhances CARMA's effectiveness.

Task complexity further differentiates CARMA's effect. CARMA's advantages align with its methodological design, particularly in tasks requiring explicit structural reinforcement. \ul{In IDM, where systematicity and substitutivity are critical, CARMA ensures structured mappings hold under perturbation, particularly in Gemma-2B (+14.6\% over the original) and Llama-1B (+2692.5\% over the original in SC)}. However, in SC, where compositionality is more distributed, larger models show lower differences between CARMA and FT, reinforcing that larger models encode sentiment shifts effectively without additional intervention.

These results strengthen the hypothesis that CARMA enhances model robustness across perturbations, particularly in structured learning tasks and models where fine-tuning alone does not fully capture compositional dependencies. While FT maintains an advantage in absolute accuracy, CARMA ensures greater consistency, making it critical for improving compositional alignment and mitigating instability in high-variance settings.

\begin{figure}[h!]
    \centering
    \includegraphics[width=.9\linewidth]{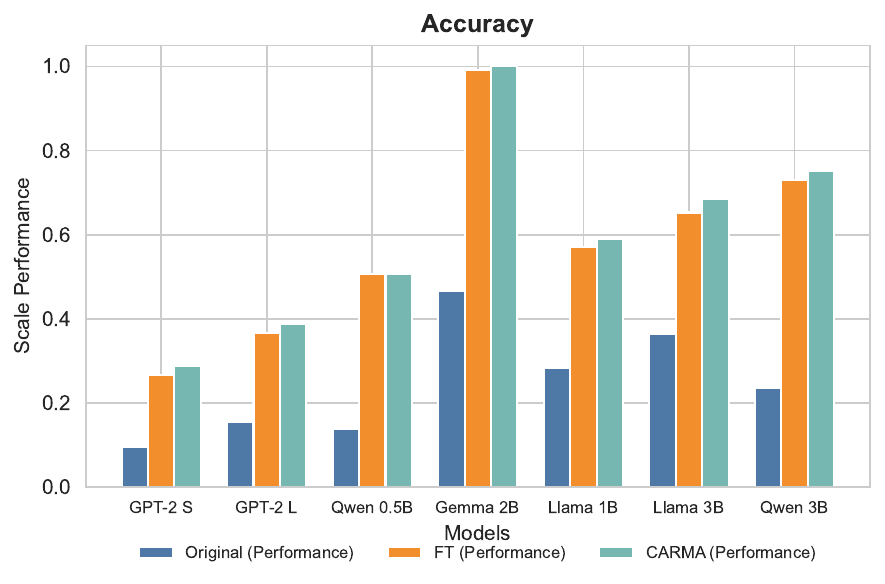}
    \caption{Task performance in IDM across GPT2 (S, L), Gemma-2B, Llama (1B, 3B), and Qwen (0.5B, 3B).}
    \label{fig:idm_performance_comparison}
\end{figure}
\begin{figure}[h!]
    \centering
    \includegraphics[width=.9\linewidth]{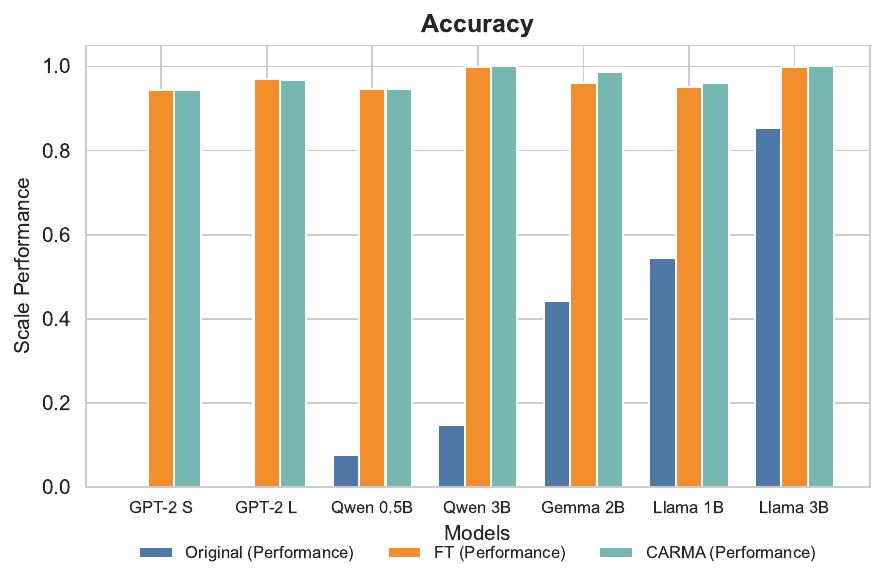}
    \caption{Task performance in SC across GPT2 (S, L), Gemma-2B, Llama (1B, 3B) and Qwen (0.5B, 3B).}
    \label{fig:sst_performance_comparison}
\end{figure}

\subsection{Impact of CARMA on Performance}\label{sec:carma_performance}
Fig.~\ref{fig:idm_performance_comparison} and~\ref{fig:sst_performance_comparison} show the performance of original, FT, and CARMA accuracies across tasks. \ul{CARMA demonstrates significant improvements over original models across tasks. For example, in IDM, GPT2-L achieves 150\% improvement, and Llama-3B shows an 89.6\% increase, while in SC, Gemma-2B demonstrates 122.5\% improvement over Original baselines}.

Task-specific patterns emerge when comparing models. For example, in IDM, CARMA outperforms FT, with Llama-3B showing a +5\% gain and GPT2-L improving by 1.7\%. In SC, CARMA maintains comparable performance to FT while enhancing robustness, suggesting it preserves learned features while strengthening compositional consistency.

\ul{CARMA enhances FT by improving representation stability and preventing feature drift, ensuring structured compositional consistency}. Its benefits are most pronounced in larger models, where greater capacity supports robust representations while maintaining fine-tuned performance. This scalability highlights CARMA's effectiveness in regularising model representations and reinforcing compositional structure without disrupting learned task features, providing a reliable solution for improving compositional reasoning in LLMs.


\section{Related work}\label{sec:related}
Research on CG in LLMs has revealed both capabilities and limitations \cite{tull2024towards, moisio-etal-2023-evaluating, sinha2024survey}, though many studies lack mechanistic analysis or concrete suggestions for improvements.

\noindent\textbf{Architectural modifications} are a common approach to tackle CG challenges. Recent proposals include pushdown layers for recursive attention \cite{murty-etal-2023-pushdown}, Layer-wise Representation Fusion for dynamic encoder weighting \cite{lin-etal-2023-learning}, and specialised semantic parsing methods \cite{shaw-etal-2021-compositional}. While effective for specific tasks, these solutions face scalability challenges due to computational overhead, specialised annotation requirements, and architectural constraints.


\noindent\textbf{Regularisation methods} provide alternative approaches through consistency regularisation \cite{yin-etal-2023-consistency}, data augmentation strategies \cite{ontanon-etal-2022-making}, and attention stability mechanisms \cite{zhai2023stabilizing}. Studies show dataset complexity and example frequency variations improve compositional reasoning \cite{zhou-etal-2023-data}. However, these methods face key limitations: token-level approaches lack adaptability to complex structures, augmentation shows diminishing returns on real data, and stability mechanisms prioritise training stability over compositional generalisation.


\noindent\textbf{Evaluation challenges} persist in CG research. Standard benchmarks like SCAN \cite{Lake2017GeneralizationWS}, PCFG \cite{ijcai2020p708}, and COGS \cite{kim-linzen-2020-cogs} rely heavily on synthetic data, limiting real-world applicability. Recent frameworks like CoGnition \cite{li-etal-2021-compositional} and CAP \cite{aljaafari2024interpreting} better align with natural language phenomena, but evaluation gaps remain. Current approaches often sacrifice generalisability for task-specific performance. \textit{CARMA} addresses these limitations through a \textit{task-agnostic, efficient solution} that enhances CG while maintaining robust cross-task performance.



\section{Conclusion}\label{sec:conclusion}
This paper presents \textsc{CARMA}, a method for enhancing compositional generalisation in LLMs through mutual information regularisation and layer-wise stability constraints. By addressing information fragmentation and instability across layers, \textsc{CARMA} improves performance robustness and stability under intervention. The method requires no architectural changes and integrates cleanly into standard fine-tuning pipelines. Future work should explore extending \textsc{CARMA} to tasks requiring more nuanced semantic reasoning and to multilingual contexts. Another important direction is combining CARMA with techniques that explicitly challenge generalisation, such as adversarial perturbations or structured distribution shifts, to promote the acquisition of novel compositional behaviours. Incorporating CARMA into improved, task-targeted architectures may further enhance its effectiveness. Additionally, controlled training-from-scratch studies could isolate CARMA’s impact more precisely and reveal deeper insights into how it shapes compositional representations across training regimes.

\section*{Limitations}
The limitations of this paper can be summed up as follows: First, our results are primarily reported for the English language. Further analysis across languages with diverse linguistic structures is left as a confirmatory future work. Second, the datasets (WordNet and SST) lack a more comprehensive representativeness of broader linguistic phenomena. Third, our focus is predominantly on decoder-based Transformers, and the employed Transformer models may inherit potential biases ingrained from their pre-training data. Finally, while CARMA maintains inference efficiency, it introduces training-time overhead due to auxiliary losses, which should be considered when integrating the method into resource-constrained environments.

\section*{Ethical statement}
This work aims to enhance language model robustness and compositional understanding through CARMA. While improving model reliability is beneficial, we acknowledge potential risks in enhancing language model capabilities. Our evaluation focuses on controlled tasks (IDM and SC) with comprehensive stability metrics to ensure responsible development and transparent reporting of model behaviour under perturbations.

\bibliography{references}

\begin{thebibliography}{55}
\providecommand{\natexlab}[1]{#1}

\bibitem[{Aljaafari et~al.(2024)Aljaafari, Carvalho, and Freitas}]{aljaafari2024interpreting}
Nura Aljaafari, Danilo~S Carvalho, and Andr{\'e} Freitas. 2024.
\newblock Interpreting token compositionality in llms: A robustness analysis.
\newblock \emph{arXiv preprint arXiv:2410.12924}.

\bibitem[{Bird et~al.(2009)Bird, Klein, and Loper}]{bird2009natural}
Steven Bird, Ewan Klein, and Edward Loper. 2009.
\newblock \emph{Natural language processing with Python: analyzing text with the natural language toolkit}.
\newblock " O'Reilly Media, Inc.".

\bibitem[{Cai et~al.(2021)Cai, Huang, Bian, and Church}]{cai2021isotropy}
Xingyu Cai, Jiaji Huang, Yuchen Bian, and Kenneth Church. 2021.
\newblock Isotropy in the contextual embedding space: Clusters and manifolds.
\newblock In \emph{International conference on learning representations}.

\bibitem[{Chi et~al.(2023)Chi, Fan, Rudnicky, and Ramadge}]{chi-etal-2023-transformer}
Ta-Chung Chi, Ting-Han Fan, Alexander Rudnicky, and Peter Ramadge. 2023.
\newblock \href {https://doi.org/10.18653/v1/2023.findings-emnlp.397} {Transformer working memory enables regular language reasoning and natural language length extrapolation}.
\newblock In \emph{Findings of the Association for Computational Linguistics: EMNLP 2023}, pages 5972--5984, Singapore. Association for Computational Linguistics.

\bibitem[{Csord{\'a}s et~al.(2021)Csord{\'a}s, Irie, and Schmidhuber}]{csordas-etal-2021-devil}
R{\'o}bert Csord{\'a}s, Kazuki Irie, and Juergen Schmidhuber. 2021.
\newblock \href {https://doi.org/10.18653/v1/2021.emnlp-main.49} {The devil is in the detail: Simple tricks improve systematic generalization of transformers}.
\newblock In \emph{Proceedings of the 2021 Conference on Empirical Methods in Natural Language Processing}, pages 619--634, Online and Punta Cana, Dominican Republic. Association for Computational Linguistics.

\bibitem[{Dankers et~al.(2022{\natexlab{a}})Dankers, Bruni, and Hupkes}]{dankers-etal-2022-paradox}
Verna Dankers, Elia Bruni, and Dieuwke Hupkes. 2022{\natexlab{a}}.
\newblock \href {https://doi.org/10.18653/v1/2022.acl-long.286} {The paradox of the compositionality of natural language: A neural machine translation case study}.
\newblock In \emph{Proceedings of the 60th Annual Meeting of the Association for Computational Linguistics (Volume 1: Long Papers)}, pages 4154--4175, Dublin, Ireland. Association for Computational Linguistics.

\bibitem[{Dankers et~al.(2022{\natexlab{b}})Dankers, Bruni, and Hupkes}]{dankers2022paradox}
Verna Dankers, Elia Bruni, and Dieuwke Hupkes. 2022{\natexlab{b}}.
\newblock \href {https://doi.org/10.18653/v1/2022.acl-long.286} {The paradox of the compositionality of natural language: A neural machine translation case study}.
\newblock In \emph{Proceedings of the 60th Annual Meeting of the Association for Computational Linguistics (Volume 1: Long Papers)}, pages 4154--4175. Association for Computational Linguistics.

\bibitem[{Dubey et~al.(2024)Dubey, Jauhri, Pandey, Kadian, Al-Dahle, Letman, Mathur, Schelten, Yang, Fan et~al.}]{dubey2024llama}
Abhimanyu Dubey, Abhinav Jauhri, Abhinav Pandey, Abhishek Kadian, Ahmad Al-Dahle, Aiesha Letman, Akhil Mathur, Alan Schelten, Amy Yang, Angela Fan, et~al. 2024.
\newblock The llama 3 herd of models.
\newblock \emph{arXiv preprint arXiv:2407.21783}.

\bibitem[{Dziri et~al.(2024)Dziri, Lu, Sclar, Li, Jiang, Lin, Welleck, West, Bhagavatula, Le~Bras et~al.}]{dziri2024faith}
Nouha Dziri, Ximing Lu, Melanie Sclar, Xiang~Lorraine Li, Liwei Jiang, Bill~Yuchen Lin, Sean Welleck, Peter West, Chandra Bhagavatula, Ronan Le~Bras, et~al. 2024.
\newblock Faith and fate: Limits of transformers on compositionality.
\newblock \emph{Advances in Neural Information Processing Systems}, 36.

\bibitem[{Fellbaum(1998)}]{fellbaum1998wordnet}
Christiane Fellbaum. 1998.
\newblock Wordnet: An electronic lexical database.
\newblock \emph{MIT Press google schola}, 2:678--686.

\bibitem[{Feucht et~al.(2024)Feucht, Atkinson, Wallace, and Bau}]{feucht2024footprints}
Sheridan Feucht, David Atkinson, Byron Wallace, and David Bau. 2024.
\newblock \href {https://arxiv.org/abs/2406.20086} {Token erasure as a footprint of implicit vocabulary items in llms}.
\newblock In \emph{The 2024 Conference on Empirical Methods in Natural Language Processing}.

\bibitem[{He et~al.(2024)He, Chen, Nie, Li, and Brennan}]{he-etal-2024-decoding}
Linyang He, Peili Chen, Ercong Nie, Yuanning Li, and Jonathan~R. Brennan. 2024.
\newblock \href {https://aclanthology.org/2024.lrec-main.402/} {Decoding probing: Revealing internal linguistic structures in neural language models using minimal pairs}.
\newblock In \emph{Proceedings of the 2024 Joint International Conference on Computational Linguistics, Language Resources and Evaluation (LREC-COLING 2024)}, pages 4488--4497, Torino, Italia. ELRA and ICCL.

\bibitem[{Honnibal et~al.(2020)Honnibal, Montani, Van~Landeghem, and Boyd}]{honnibal2020spacy}
Matthew Honnibal, Ines Montani, Sofie Van~Landeghem, and Adriane Boyd. 2020.
\newblock \href {https://doi.org/10.5281/zenodo.1212303} {spacy: Industrial-strength natural language processing in python}.

\bibitem[{Hupkes et~al.(2020)Hupkes, Dankers, Mul, and Bruni}]{ijcai2020p708}
Dieuwke Hupkes, Verna Dankers, Mathijs Mul, and Elia Bruni. 2020.
\newblock \href {https://doi.org/10.24963/ijcai.2020/708} {Compositionality decomposed: How do neural networks generalise? (extended abstract)}.
\newblock In \emph{Proceedings of the Twenty-Ninth International Joint Conference on Artificial Intelligence, {IJCAI-20}}, pages 5065--5069. International Joint Conferences on Artificial Intelligence Organization.
\newblock Journal track.

\bibitem[{Ismayilzada et~al.(2024)Ismayilzada, Circi, S{\"a}lev{\"a}, Sirin, K{\"o}ksal, Dhingra, Bosselut, van~der Plas, and Ataman}]{ismayilzada2024evaluating}
Mete Ismayilzada, Defne Circi, Jonne S{\"a}lev{\"a}, Hale Sirin, Abdullatif K{\"o}ksal, Bhuwan Dhingra, Antoine Bosselut, Lonneke van~der Plas, and Duygu Ataman. 2024.
\newblock Evaluating morphological compositional generalization in large language models.
\newblock \emph{arXiv preprint arXiv:2410.12656}.

\bibitem[{Kim and Linzen(2020{\natexlab{a}})}]{kim2020cogs}
Najoung Kim and Tal Linzen. 2020{\natexlab{a}}.
\newblock Cogs: A compositional generalization challenge based on semantic interpretation.
\newblock In \emph{Proceedings of the 2020 conference on empirical methods in natural language processing (emnlp)}, pages 9087--9105.

\bibitem[{Kim and Linzen(2020{\natexlab{b}})}]{kim-linzen-2020-cogs}
Najoung Kim and Tal Linzen. 2020{\natexlab{b}}.
\newblock \href {https://doi.org/10.18653/v1/2020.emnlp-main.731} {{COGS}: A compositional generalization challenge based on semantic interpretation}.
\newblock In \emph{Proceedings of the 2020 Conference on Empirical Methods in Natural Language Processing (EMNLP)}, pages 9087--9105, Online. Association for Computational Linguistics.

\bibitem[{Kitaev et~al.(2019)Kitaev, Cao, and Klein}]{kitaev-etal-2019-multilingual}
Nikita Kitaev, Steven Cao, and Dan Klein. 2019.
\newblock \href {https://doi.org/10.18653/v1/P19-1340} {Multilingual constituency parsing with self-attention and pre-training}.
\newblock In \emph{Proceedings of the 57th Annual Meeting of the Association for Computational Linguistics}, pages 3499--3505, Florence, Italy. Association for Computational Linguistics.

\bibitem[{Kitaev and Klein(2018)}]{kitaev-klein-2018-constituency}
Nikita Kitaev and Dan Klein. 2018.
\newblock \href {https://doi.org/10.18653/v1/P18-1249} {Constituency parsing with a self-attentive encoder}.
\newblock In \emph{Proceedings of the 56th Annual Meeting of the Association for Computational Linguistics (Volume 1: Long Papers)}, pages 2676--2686, Melbourne, Australia. Association for Computational Linguistics.

\bibitem[{Lake and Baroni(2017)}]{Lake2017GeneralizationWS}
Brenden~M. Lake and Marco Baroni. 2017.
\newblock \href {https://api.semanticscholar.org/CorpusID:46761158} {Generalization without systematicity: On the compositional skills of sequence-to-sequence recurrent networks}.
\newblock In \emph{International Conference on Machine Learning}.

\bibitem[{Langedijk et~al.(2024)Langedijk, Mohebbi, Sarti, Zuidema, and Jumelet}]{langedijk-etal-2024-decoderlens}
Anna Langedijk, Hosein Mohebbi, Gabriele Sarti, Willem Zuidema, and Jaap Jumelet. 2024.
\newblock \href {https://doi.org/10.18653/v1/2024.findings-naacl.296} {{D}ecoder{L}ens: Layerwise interpretation of encoder-decoder transformers}.
\newblock In \emph{Findings of the Association for Computational Linguistics: NAACL 2024}, pages 4764--4780, Mexico City, Mexico. Association for Computational Linguistics.

\bibitem[{Li et~al.(2021)Li, Yin, Chen, and Zhang}]{li-etal-2021-compositional}
Yafu Li, Yongjing Yin, Yulong Chen, and Yue Zhang. 2021.
\newblock \href {https://doi.org/10.18653/v1/2021.acl-long.368} {On compositional generalization of neural machine translation}.
\newblock In \emph{Proceedings of the 59th Annual Meeting of the Association for Computational Linguistics and the 11th International Joint Conference on Natural Language Processing (Volume 1: Long Papers)}, pages 4767--4780, Online. Association for Computational Linguistics.

\bibitem[{Li et~al.(2023)Li, Wei, and Lian}]{li-etal-2023-learning}
Zhaoyi Li, Ying Wei, and Defu Lian. 2023.
\newblock \href {https://doi.org/10.18653/v1/2023.acl-long.157} {Learning to substitute spans towards improving compositional generalization}.
\newblock In \emph{Proceedings of the 61st Annual Meeting of the Association for Computational Linguistics (Volume 1: Long Papers)}, pages 2791--2811, Toronto, Canada. Association for Computational Linguistics.

\bibitem[{Lin et~al.(2023)Lin, Li, Zheng, Fu, Liu, Chen, and Shi}]{lin-etal-2023-learning}
Lei Lin, Shuangtao Li, Yafang Zheng, Biao Fu, Shan Liu, Yidong Chen, and Xiaodong Shi. 2023.
\newblock \href {https://doi.org/10.18653/v1/2023.findings-emnlp.108} {Learning to compose representations of different encoder layers towards improving compositional generalization}.
\newblock In \emph{Findings of the Association for Computational Linguistics: EMNLP 2023}, pages 1599--1614, Singapore. Association for Computational Linguistics.

\bibitem[{Loria et~al.()}]{loria2018textblob}
Steven Loria et~al.
\newblock textblob documentation.
\newblock \emph{Release 0.18.0}.

\bibitem[{Miller(1994)}]{miller-1994-wordnet}
George~A. Miller. 1994.
\newblock \href {https://aclanthology.org/H94-1111} {{W}ord{N}et: A lexical database for {E}nglish}.
\newblock In \emph{{H}uman {L}anguage {T}echnology: Proceedings of a Workshop held at {P}lainsboro, {N}ew {J}ersey, {M}arch 8-11, 1994}.

\bibitem[{Moisio et~al.(2023)Moisio, Creutz, and Kurimo}]{moisio-etal-2023-evaluating}
Anssi Moisio, Mathias Creutz, and Mikko Kurimo. 2023.
\newblock \href {https://aclanthology.org/2023.nodalida-1.75} {Evaluating morphological generalisation in machine translation by distribution-based compositionality assessment}.
\newblock In \emph{Proceedings of the 24th Nordic Conference on Computational Linguistics (NoDaLiDa)}, pages 738--751, T{\'o}rshavn, Faroe Islands. University of Tartu Library.

\bibitem[{Murty et~al.(2023)Murty, Sharma, Andreas, and Manning}]{murty-etal-2023-pushdown}
Shikhar Murty, Pratyusha Sharma, Jacob Andreas, and Christopher Manning. 2023.
\newblock \href {https://doi.org/10.18653/v1/2023.emnlp-main.195} {Pushdown layers: Encoding recursive structure in transformer language models}.
\newblock In \emph{Proceedings of the 2023 Conference on Empirical Methods in Natural Language Processing}, pages 3233--3247, Singapore. Association for Computational Linguistics.

\bibitem[{Nanda and Bloom(2022)}]{nanda2022transformerlens}
Neel Nanda and Joseph Bloom. 2022.
\newblock Transformerlens.
\newblock \url{https://github.com/TransformerLensOrg/TransformerLens}.

\bibitem[{Ontanon et~al.(2022)Ontanon, Ainslie, Fisher, and Cvicek}]{ontanon-etal-2022-making}
Santiago Ontanon, Joshua Ainslie, Zachary Fisher, and Vaclav Cvicek. 2022.
\newblock \href {https://doi.org/10.18653/v1/2022.acl-long.251} {Making transformers solve compositional tasks}.
\newblock In \emph{Proceedings of the 60th Annual Meeting of the Association for Computational Linguistics (Volume 1: Long Papers)}, pages 3591--3607, Dublin, Ireland. Association for Computational Linguistics.

\bibitem[{Oord et~al.(2018)Oord, Li, and Vinyals}]{oord2018representation}
Aaron van~den Oord, Yazhe Li, and Oriol Vinyals. 2018.
\newblock Representation learning with contrastive predictive coding.
\newblock \emph{arXiv preprint arXiv:1807.03748}.

\bibitem[{Partee(1984)}]{Partee1984}
Barbara~H. Partee. 1984.
\newblock Compositionality.
\newblock In Fred Landman and Frank Veltman, editors, \emph{Varieties of Formal Semantics}, pages 281--312. Foris Publications.

\bibitem[{Paszke et~al.(2019)Paszke, Gross, Massa, Lerer, Bradbury, Chanan, Killeen, Lin, Gimelshein, Antiga et~al.}]{paszke2019pytorch}
Adam Paszke, Sam Gross, Francisco Massa, Adam Lerer, James Bradbury, Gregory Chanan, Trevor Killeen, Zeming Lin, Natalia Gimelshein, Luca Antiga, et~al. 2019.
\newblock Pytorch: An imperative style, high-performance deep learning library.
\newblock \emph{Advances in neural information processing systems}, 32.

\bibitem[{Pedregosa et~al.(2011)Pedregosa, Varoquaux, Gramfort, Michel, Thirion, Grisel, Blondel, Prettenhofer, Weiss, Dubourg, Vanderplas, Passos, Cournapeau, Brucher, Perrot, and Duchesnay}]{scikit-learn}
F.~Pedregosa, G.~Varoquaux, A.~Gramfort, V.~Michel, B.~Thirion, O.~Grisel, M.~Blondel, P.~Prettenhofer, R.~Weiss, V.~Dubourg, J.~Vanderplas, A.~Passos, D.~Cournapeau, M.~Brucher, M.~Perrot, and E.~Duchesnay. 2011.
\newblock Scikit-learn: Machine learning in {P}ython.
\newblock \emph{Journal of Machine Learning Research}, 12:2825--2830.

\bibitem[{Petty et~al.(2024)Petty, Steenkiste, Dasgupta, Sha, Garrette, and Linzen}]{petty2024impact}
Jackson Petty, Sjoerd Steenkiste, Ishita Dasgupta, Fei Sha, Dan Garrette, and Tal Linzen. 2024.
\newblock The impact of depth on compositional generalization in transformer language models.
\newblock In \emph{Proceedings of the 2024 Conference of the North American Chapter of the Association for Computational Linguistics: Human Language Technologies (Volume 1: Long Papers)}, pages 7232--7245.

\bibitem[{Radford et~al.(2019)Radford, Wu, Child, Luan, Amodei, Sutskever et~al.}]{radford2019language}
Alec Radford, Jeffrey Wu, Rewon Child, David Luan, Dario Amodei, Ilya Sutskever, et~al. 2019.
\newblock Language models are unsupervised multitask learners.
\newblock \emph{OpenAI blog}, 1(8):9.

\bibitem[{Ram et~al.(2024)Ram, Klinger, and Gray}]{ram2024makes}
Parikshit Ram, Tim Klinger, and Alexander Gray. 2024.
\newblock \href {https://doi.org/10.24963/ijcai.2024/533} {What makes models compositional? a theoretical view}.
\newblock In \emph{Proceedings of the Thirty-Third International Joint Conference on Artificial Intelligence ({IJCAI}-24)}, pages 4824--4832. International Joint Conferences on Artificial Intelligence Organization.

\bibitem[{Ramesh et~al.(2024)Ramesh, Lubana, Khona, Dick, and Tanaka}]{ramesh2024compositional}
Rahul Ramesh, Ekdeep~Singh Lubana, Mikail Khona, Robert~P. Dick, and Hidenori Tanaka. 2024.
\newblock \href {https://openreview.net/forum?id=L1eJ3NKPCd} {Compositional capabilities of autoregressive transformers: A study on synthetic, interpretable tasks}.
\newblock In \emph{Forty-first International Conference on Machine Learning}.

\bibitem[{Sajjadi et~al.(2016)Sajjadi, Javanmardi, and Tasdizen}]{sajjadi2016regularization}
Mehdi Sajjadi, Mehran Javanmardi, and Tolga Tasdizen. 2016.
\newblock Regularization with stochastic transformations and perturbations for deep semi-supervised learning.
\newblock \emph{Advances in neural information processing systems}, 29.

\bibitem[{Shaw et~al.(2021)Shaw, Chang, Pasupat, and Toutanova}]{shaw-etal-2021-compositional}
Peter Shaw, Ming-Wei Chang, Panupong Pasupat, and Kristina Toutanova. 2021.
\newblock \href {https://doi.org/10.18653/v1/2021.acl-long.75} {Compositional generalization and natural language variation: Can a semantic parsing approach handle both?}
\newblock In \emph{Proceedings of the 59th Annual Meeting of the Association for Computational Linguistics and the 11th International Joint Conference on Natural Language Processing (Volume 1: Long Papers)}, pages 922--938, Online. Association for Computational Linguistics.

\bibitem[{Sikarwar et~al.(2022)Sikarwar, Patel, and Goyal}]{sikarwar-etal-2022-transformers}
Ankur Sikarwar, Arkil Patel, and Navin Goyal. 2022.
\newblock \href {https://doi.org/10.18653/v1/2022.emnlp-main.41} {When can transformers ground and compose: Insights from compositional generalization benchmarks}.
\newblock In \emph{Proceedings of the 2022 Conference on Empirical Methods in Natural Language Processing}, pages 648--669, Abu Dhabi, United Arab Emirates. Association for Computational Linguistics.

\bibitem[{Sinha et~al.(2024)Sinha, Premsri, and Kordjamshidi}]{sinha2024survey}
Sania Sinha, Tanawan Premsri, and Parisa Kordjamshidi. 2024.
\newblock A survey on compositional learning of ai models: Theoretical and experimetnal practices.
\newblock \emph{arXiv preprint arXiv:2406.08787}.

\bibitem[{Socher et~al.(2013)Socher, Perelygin, Wu, Chuang, Manning, Ng, and Potts}]{socher2013recursive}
Richard Socher, Alex Perelygin, Jean Wu, Jason Chuang, Christopher~D Manning, Andrew~Y Ng, and Christopher Potts. 2013.
\newblock Recursive deep models for semantic compositionality over a sentiment treebank.
\newblock In \emph{Proceedings of the 2013 conference on empirical methods in natural language processing}, pages 1631--1642.

\bibitem[{Team et~al.(2024)Team, Riviere, Pathak, Sessa, Hardin, Bhupatiraju, Hussenot, Mesnard, Shahriari, Ram{\'e} et~al.}]{team2024gemma}
Gemma Team, Morgane Riviere, Shreya Pathak, Pier~Giuseppe Sessa, Cassidy Hardin, Surya Bhupatiraju, L{\'e}onard Hussenot, Thomas Mesnard, Bobak Shahriari, Alexandre Ram{\'e}, et~al. 2024.
\newblock Gemma 2: Improving open language models at a practical size.
\newblock \emph{arXiv preprint arXiv:2408.00118}.

\bibitem[{Tull et~al.(2024)Tull, Lorenz, Clark, Khan, and Coecke}]{tull2024towards}
Sean Tull, Robin Lorenz, Stephen Clark, Ilyas Khan, and Bob Coecke. 2024.
\newblock Towards compositional interpretability for xai.
\newblock \emph{arXiv preprint arXiv:2406.17583}.

\bibitem[{Wolf et~al.(2019)Wolf, Debut, Sanh, Chaumond, Delangue, Moi, Cistac, Rault, Louf, Funtowicz, Davison, Shleifer, von Platen, Ma, Jernite, Plu, Xu, Scao, Gugger, Drame, Lhoest, and Rush}]{wolf2019huggingface}
Thomas Wolf, Lysandre Debut, Victor Sanh, Julien Chaumond, Clement Delangue, Anthony Moi, Pierric Cistac, Tim Rault, R{\'e}mi Louf, Morgan Funtowicz, Joe Davison, Sam Shleifer, Patrick von Platen, Clara Ma, Yacine Jernite, Julien Plu, Canwen Xu, Teven~Le Scao, Sylvain Gugger, Mariama Drame, Quentin Lhoest, and Alexander Rush. 2019.
\newblock Huggingface's transformers: State-of-the-art natural language processing.
\newblock \emph{arXiv preprint arXiv:1910.03771}.

\bibitem[{Wolf et~al.(2020)Wolf, Debut, Sanh, Chaumond, Delangue, Moi, Cistac, Rault, Louf, Funtowicz, Davison, Shleifer, von Platen, Ma, Jernite, Plu, Xu, Scao, Gugger, Drame, Lhoest, and Rush}]{wolf-etal-2020-transformers}
Thomas Wolf, Lysandre Debut, Victor Sanh, Julien Chaumond, Clement Delangue, Anthony Moi, Pierric Cistac, Tim Rault, Rémi Louf, Morgan Funtowicz, Joe Davison, Sam Shleifer, Patrick von Platen, Clara Ma, Yacine Jernite, Julien Plu, Canwen Xu, Teven~Le Scao, Sylvain Gugger, Mariama Drame, Quentin Lhoest, and Alexander~M. Rush. 2020.
\newblock \href {https://www.aclweb.org/anthology/2020.emnlp-demos.6} {Transformers: State-of-the-art natural language processing}.
\newblock In \emph{Proceedings of the 2020 Conference on Empirical Methods in Natural Language Processing: System Demonstrations}, pages 38--45, Online. Association for Computational Linguistics.

\bibitem[{Yang et~al.(2024{\natexlab{a}})Yang, Yang, Zhang, Hui, Zheng, Yu, Li, Liu, Huang, Wei et~al.}]{yang2024qwen2}
An~Yang, Baosong Yang, Beichen Zhang, Binyuan Hui, Bo~Zheng, Bowen Yu, Chengyuan Li, Dayiheng Liu, Fei Huang, Haoran Wei, et~al. 2024{\natexlab{a}}.
\newblock Qwen2.5 technical report.
\newblock \emph{arXiv preprint arXiv:2412.15115}.

\bibitem[{Yang et~al.(2024{\natexlab{b}})Yang, Lu, Lam, and Cai}]{yang2024exploring}
Haoran Yang, Hongyuan Lu, Wai Lam, and Deng Cai. 2024{\natexlab{b}}.
\newblock Exploring compositional generalization of large language models.
\newblock In \emph{Proceedings of the 2024 Conference of the North American Chapter of the Association for Computational Linguistics: Human Language Technologies (Volume 4: Student Research Workshop)}, pages 16--24.

\bibitem[{Yin et~al.(2023{\natexlab{a}})Yin, Zeng, Li, Meng, Zhou, and Zhang}]{yin2023consistency}
Yongjing Yin, Jiali Zeng, Yafu Li, Fandong Meng, Jie Zhou, and Yue Zhang. 2023{\natexlab{a}}.
\newblock Consistency regularization training for compositional generalization.
\newblock In \emph{Proceedings of the 61st Annual Meeting of the Association for Computational Linguistics (Volume 1: Long Papers)}, pages 1294--1308.

\bibitem[{Yin et~al.(2023{\natexlab{b}})Yin, Zeng, Li, Meng, Zhou, and Zhang}]{yin-etal-2023-consistency}
Yongjing Yin, Jiali Zeng, Yafu Li, Fandong Meng, Jie Zhou, and Yue Zhang. 2023{\natexlab{b}}.
\newblock \href {https://doi.org/10.18653/v1/2023.acl-long.72} {Consistency regularization training for compositional generalization}.
\newblock In \emph{Proceedings of the 61st Annual Meeting of the Association for Computational Linguistics (Volume 1: Long Papers)}, pages 1294--1308, Toronto, Canada. Association for Computational Linguistics.

\bibitem[{Zhai et~al.(2023)Zhai, Likhomanenko, Littwin, Busbridge, Ramapuram, Zhang, Gu, and Susskind}]{zhai2023stabilizing}
Shuangfei Zhai, Tatiana Likhomanenko, Etai Littwin, Dan Busbridge, Jason Ramapuram, Yizhe Zhang, Jiatao Gu, and Joshua~M Susskind. 2023.
\newblock Stabilizing transformer training by preventing attention entropy collapse.
\newblock In \emph{International Conference on Machine Learning}, pages 40770--40803. PMLR.

\bibitem[{Zhang et~al.(2024)Zhang, He, Lei, Yue, Wang, and Lu}]{zhang2024can}
Min Zhang, Jianfeng He, Shuo Lei, Murong Yue, Linhan Wang, and Chang-Tien Lu. 2024.
\newblock Can llm find the green circle? investigation and human-guided tool manipulation for compositional generalization.
\newblock In \emph{ICASSP 2024-2024 IEEE International Conference on Acoustics, Speech and Signal Processing (ICASSP)}, pages 11996--12000. IEEE.

\bibitem[{Zhong et~al.(2024)Zhong, Li, Wang, Song, Wei, Lian, and Mao}]{zhong2024benchmarking}
Tianqi Zhong, Zhaoyi Li, Quan Wang, Linqi Song, Ying Wei, Defu Lian, and Zhendong Mao. 2024.
\newblock \href {https://doi.org/10.18653/v1/2024.acl-long.351} {Benchmarking and improving compositional generalization of multi-aspect controllable text generation}.
\newblock In \emph{Proceedings of the 62nd Annual Meeting of the Association for Computational Linguistics (Volume 1: Long Papers)}, pages 6486--6517. Association for Computational Linguistics.

\bibitem[{Zhou et~al.(2023)Zhou, Jiang, and Bansal}]{zhou-etal-2023-data}
Xiang Zhou, Yichen Jiang, and Mohit Bansal. 2023.
\newblock \href {https://doi.org/10.18653/v1/2023.emnlp-main.898} {Data factors for better compositional generalization}.
\newblock In \emph{Proceedings of the 2023 Conference on Empirical Methods in Natural Language Processing}, pages 14549--14566, Singapore. Association for Computational Linguistics.

\end{thebibliography}

\appendix
\section{Task Selection and Compositionality Considerations}\label{sec:task_selection}
To assess compositional generalisation and the benefits of CARMA, we targeted tasks that involve systematic meaning construction and sensitivity to structural modifications. To that end, we opted to employ Inverse Dictionary Modelling (IDM) and Sentiment Classification (SC) as proxies for different dimensions of compositionality, capturing both structured composition and hierarchical generalisation.

IDM requires models to generate a single-word representation from a natural language definition, mapping from the composition of input constituents (individual concept components) to a specific term. On the other hand, SC maps meaning to a sentiment label, aggregating local meaning elements into a global interpretation. While IDM focuses on explicit compositional mapping, SC evaluates distributed composition, where sentiment is shaped by multiple interacting components. 

\begin{figure}[h]
    \centering
    \includegraphics[width=\linewidth]{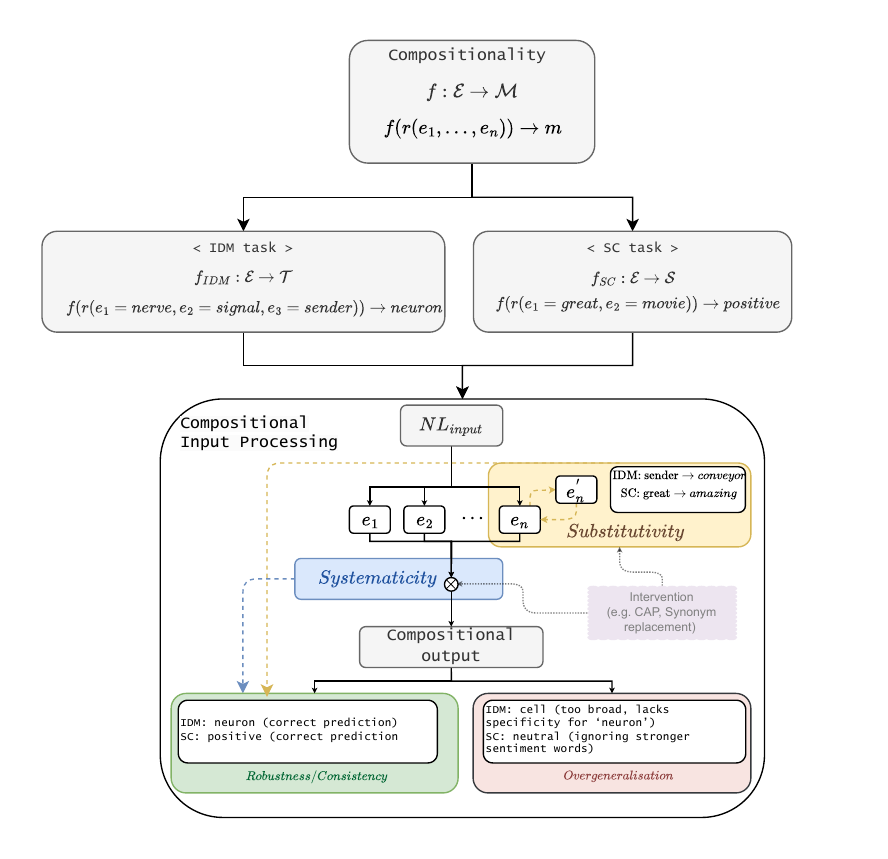}
    \caption{Illustration of compositional generalisation in Inverse Dictionary Modelling (IDM) and Sentiment Classification (SC). The figure highlights key compositional properties: systematicity ensures coherent meaning construction, substitutivity maintains meaning under lexical variations, robustness preserves intended outputs under perturbations, and over-generalisation leads to overly broad or semantically weak predictions (e.g., neuron misclassified as cell or positive reduced to neutral).}
    \label{fig:task_comps}
\end{figure}

Both tasks assess several aspects of compositionality (Figure~\ref{fig:task_comps}), namely systematicity (structured meaning formation), substitutivity (semantic preservation under transformation), and resistance to over-generalisation (ensuring bounded semantic deviation). Further, they evaluate robustness, testing whether models can maintain correctness and consistency under internal and input-lexical perturbations. IDM and SC provide a comprehensive test of compositional generalisation across structured and distributed representations.

\section{Detailed Experimental Configuration}
\subsection{Task Formalisation}\label{sec:task_formalisation}
This paper evaluates the effectiveness of CARMA in enhancing the compositional generalisation of large language models (LLMs) through two tasks. These tasks were selected based on their focus on input token structure and compositional semantics, utilising next-token prediction with single-token outputs. Formal definitions for each task are presented below.
\begin{figure*}
    \centering
    \subfigure[GPT2-S]{
        \includegraphics[width=0.48\textwidth]{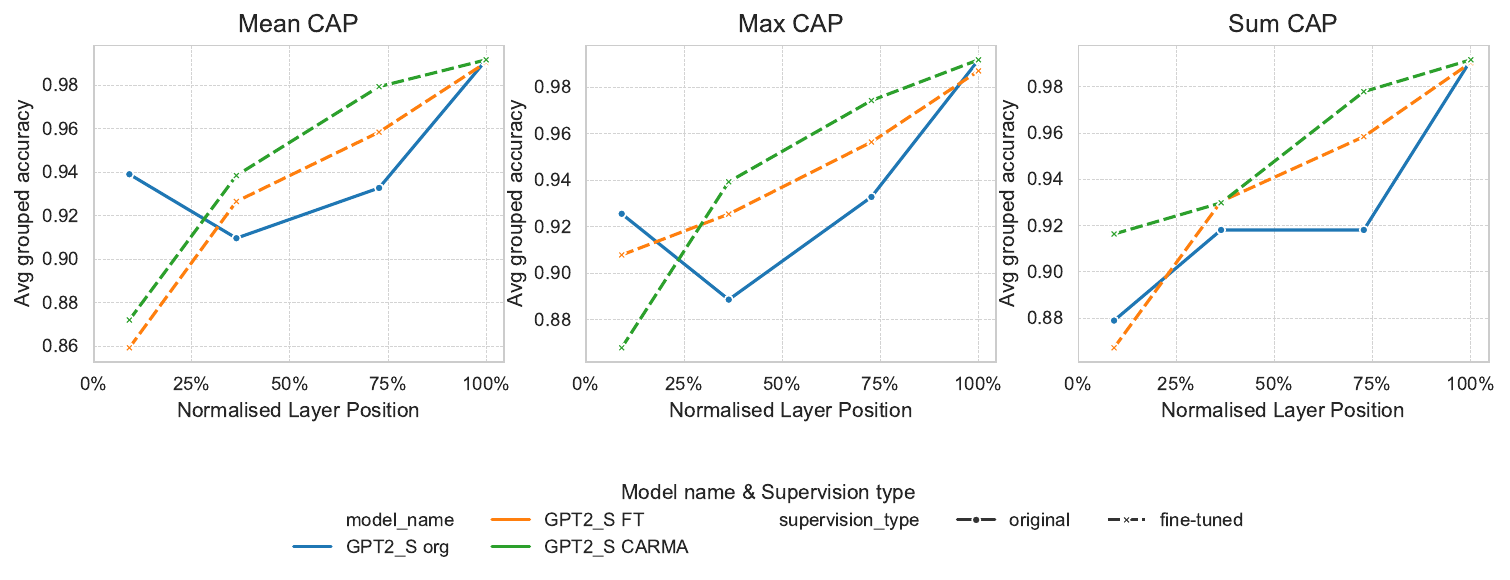}
    }
    \subfigure[GPT2-L]{
        \includegraphics[width=0.48\textwidth]{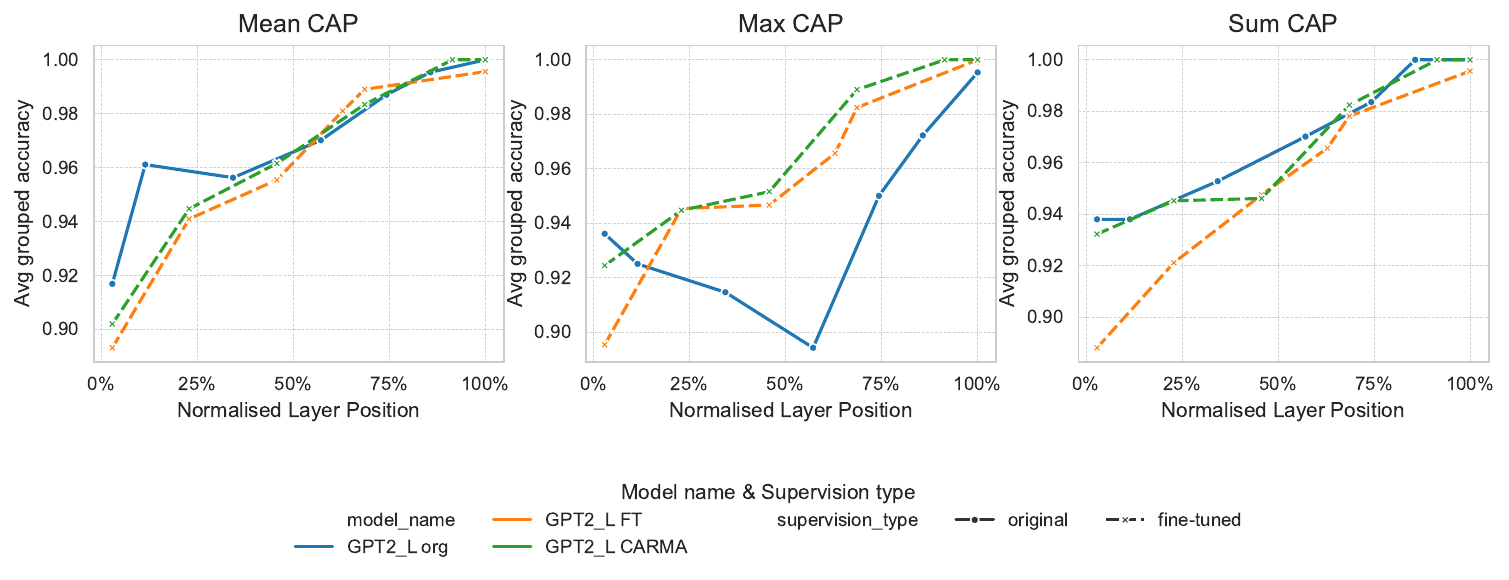}
    } \\
    \subfigure[Gemma-2B]{
        \includegraphics[width=0.48\textwidth]{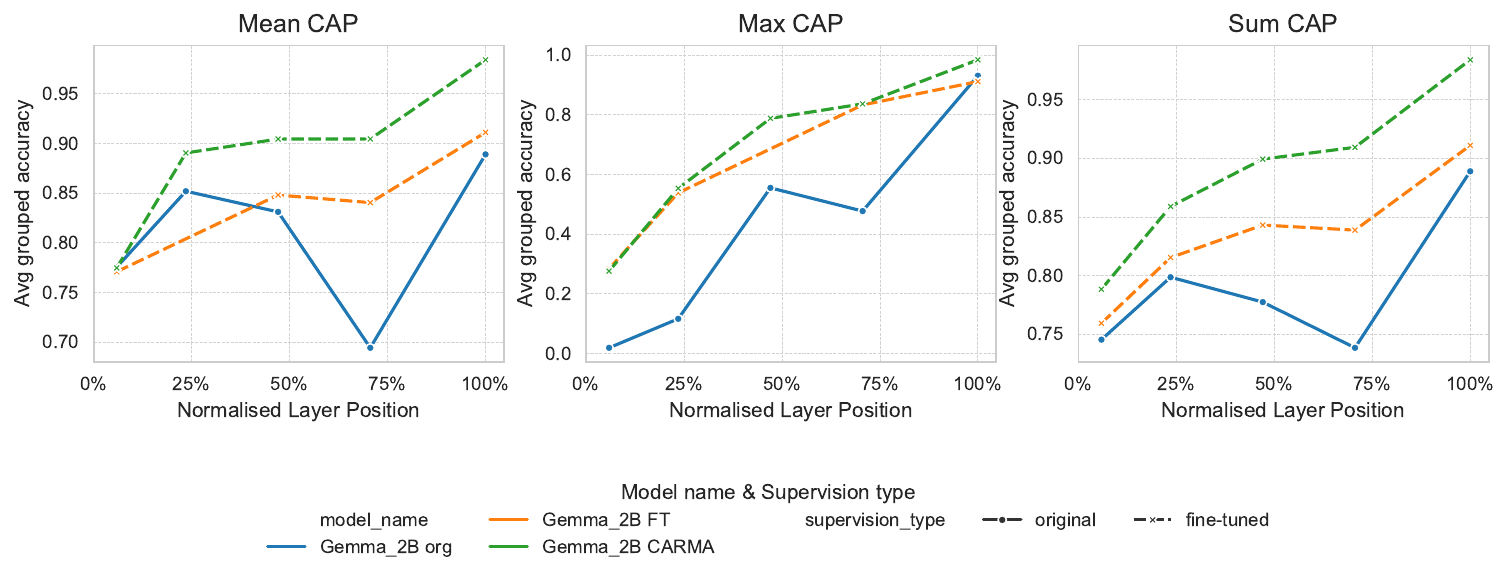}
    }
    \subfigure[Qwen-0.5B]{
        \includegraphics[width=0.48\textwidth]{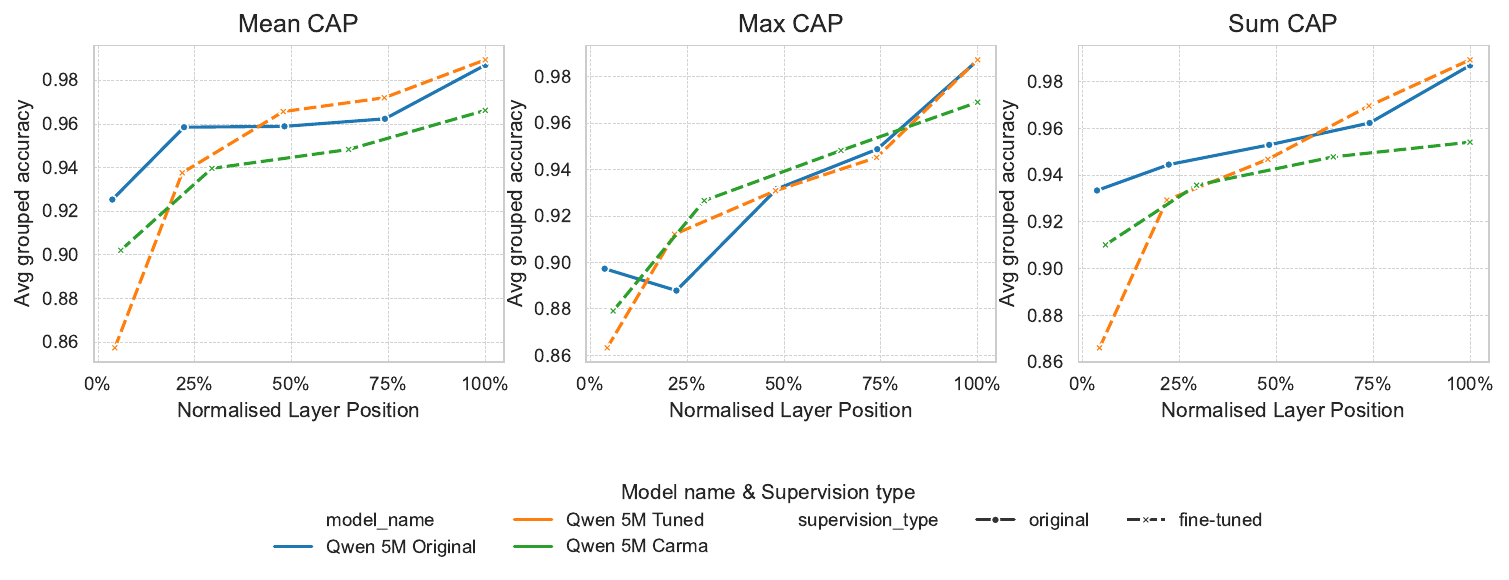}
    } \\
    \subfigure[Llama-1B]{
        \includegraphics[width=0.48\textwidth]{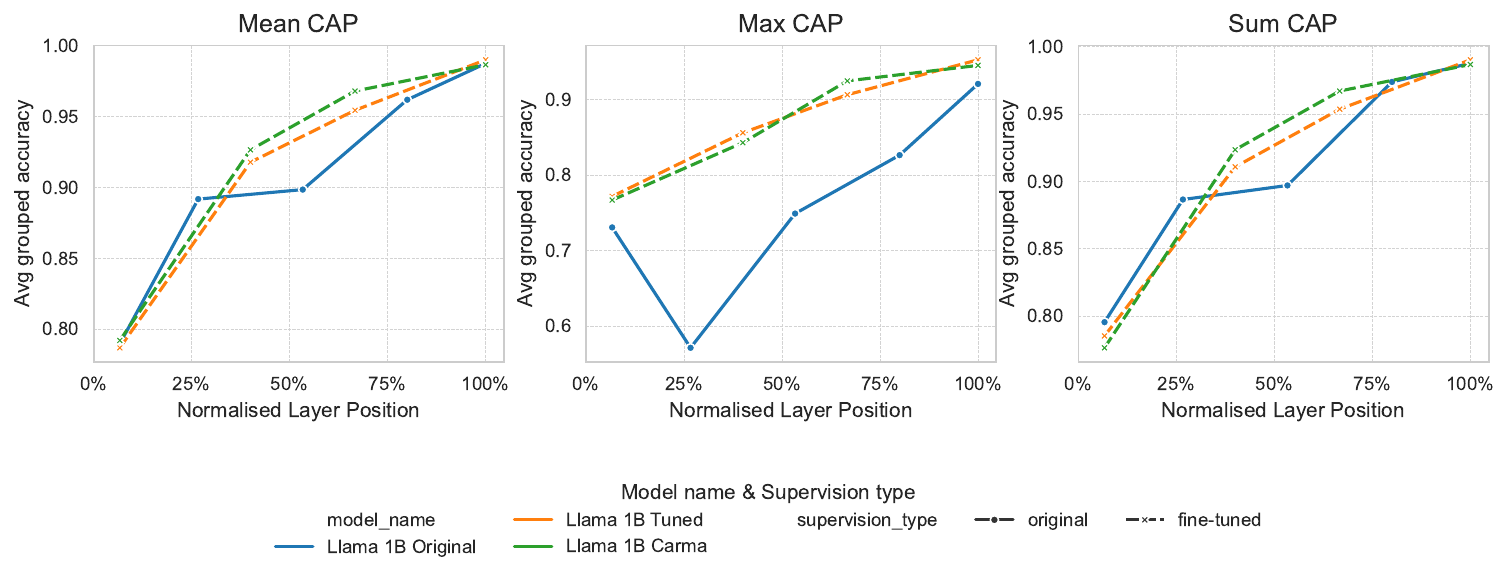}
    }
    \subfigure[Llama-3B]{
        \includegraphics[width=0.48\textwidth]{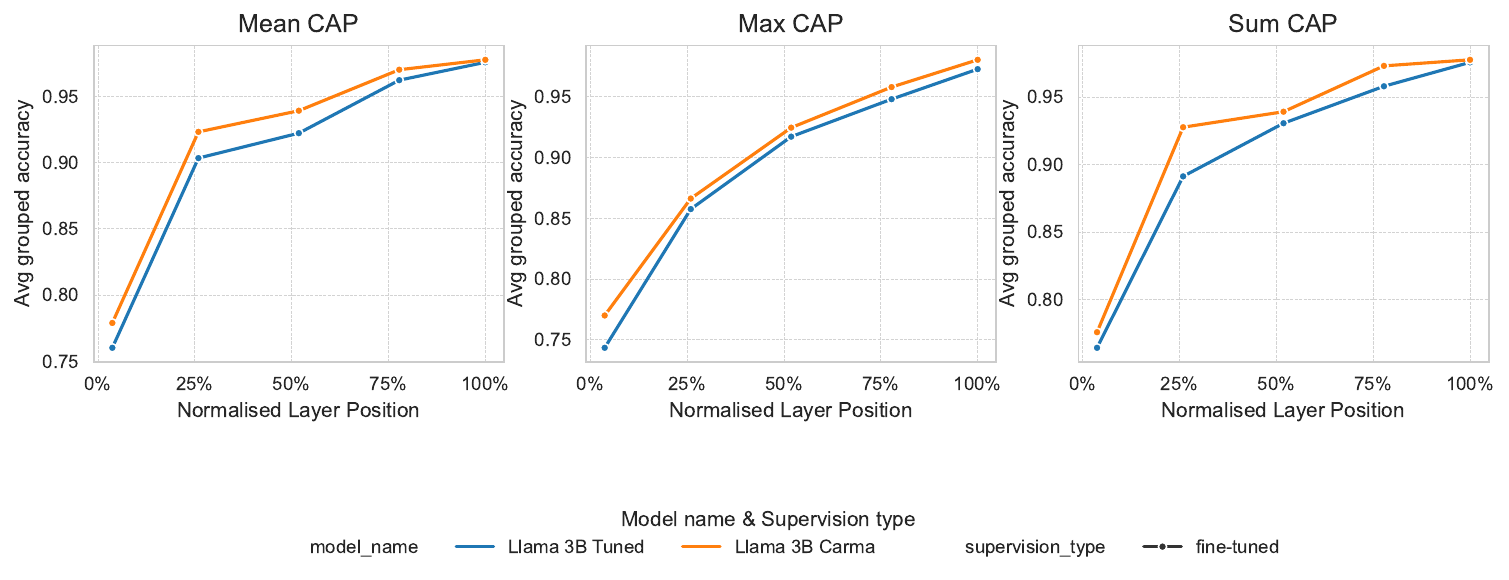}
    }
    \caption{IDM Performance Across Models Under CAP}
    \label{fig:idm_models}
\end{figure*}

\noindent\paragraph{Inverse Definition Modelling (IDM).}  
This task requires the model to predict a definiendum \(D\), given its corresponding definition \(\text{definition}\) in natural language. Formally, the definition is represented as a sequence of tokens, \(\text{definition} = \{\text{tok}_1, \text{tok}_2, \dots, \text{tok}_n\}\), and the model seeks to produce \(D\) such that:  
\begin{equation}
    D = \arg\max_{t \in \mathcal{V}} P(d \mid \text{definition}),
\end{equation} 
where \(\mathcal{V}\) denotes the model's vocabulary, and \(d\) represents a potential definiendum. Predictions are deemed correct only if they exactly match the target output.

\paragraph{Sentiment classification (SC).} 
This task involves assigning a sentiment label to a given sentence containing sentiment cues and potential modifiers. The model processes the input \(\text{sentence}\), represented as a sequence of tokens \(\text{sentence} = \{\text{tok}_1, \text{tok}_2, \dots, \text{tok}_n\}\), and produces an output \(\text{label}\) from a predefined set of sentiment classes \(\mathcal{A}\) (i.e., \textit{positive}, \textit{negative}, \textit{neutral}). Formally, the task is defined as:  
\begin{equation}
\text{label} = \arg\max_{\ell \in \mathcal{L}} P(\ell \mid \text{sentence}),
\end{equation}  
where \(P(\ell \mid \text{sentence})\) is the probability of the sentiment label \(\ell\) given the sentence. The model's performance is evaluated based on its ability to correctly predict the sentiment, accounting for compositional nuances such as modifiers and contrasts.  

\subsection{Datasets specification and pre-processing}\label{sec:dataset_appendix}
For IDM, the training and test datasets were derived from WordNet \cite{fellbaum1998wordnet}, a widely used lexical database of the English language. WordNet comprises over 117,000 synsets, each representing a distinct concept and annotated with semantic relationships such as hypernyms, synonyms, and definitions. To ensure consistency and improve data quality, standard preprocessing techniques were applied, including the removal of special characters, punctuation, extra spaces, and parenthesised content where necessary. The dataset focuses on general-purpose vocabulary rather than specialised domains or demographic groups. The dataset was initially split into an 80-20 ratio, with 80\% allocated for training. The remaining 20\% was further divided equally into validation and test sets. 

The SC dataset was derived from the Stanford Sentiment Treebank (SST) \cite{socher2013recursive}, a corpus of English movie reviews annotated for analysis of the compositional effects of sentiment inference and was released under Apache License, Version 2.0. SST includes fine-grained sentiment labels at both the phrase and sentence levels, making it a standard benchmark for evaluating sentiment classification models. The original dataset splits provided by the authors were maintained to ensure consistency in training, validation, and testing. For SST labels, sentiment scores were categorised as follows: values equal to or greater than 0.6 were classified as positive, scores between 0 and 0.6 were considered neutral, and scores below zero were assigned as negative. The final test dataset sizes for each task are presented in Table~\ref{tab:test_set_sizes}.

\begin{table}[h]
\centering
\small
\begin{tabular}{|l|c|c|c|}
\hline
\textbf{Dataset} & \textbf{Train size} & \textbf{validation Size} & \textbf{Test Size} \\
\hline
WordNet & 9563 & 1154 & 1231  \\

\hline
SST & 8544 & 1101 & 2210 \\
\hline
\end{tabular}
\caption{Train, validation, and test set sizes for WordNet and SST datasets used in this paper.}
\label{tab:test_set_sizes}
\end{table}

\subsection{Model training and fine-tuning settings}\label{sec:fine_tuning_appenix}
Table \ref{tab:model_properties} summarises the key characteristics of the models evaluated in this study. All models were obtained from Hugging Face \cite{wolf2019huggingface} under their respective licenses: GPT-2 (Modified MIT), Llama 3.2 (Meta Llama 3 Community), Qwen 2.5 (Apache 2.0), and Gemma-2B (Gemma Terms of Use). While all models were pre-trained on English data, LLama and Qwen models provide additional multilingual capabilities, namely English, German, French, Italian, Portuguese, Hindi, Spanish, and Thai for LLama, and over 10 languages, including Chinese, English, French, Spanish, Portuguese, Russian, Arabic, Japanese, Korean, Vietnamese, Thai, and Indonesian for Qwen. The models employ the following tokenisation approaches: GPT-2, Byte Pair Encoding (BPE) with a 50,257-token vocabulary, optimised primarily for English, Llama 3.2 uses SentencePiece-based BPE, combining 100K tokens from Tiktoken3 with 28K additional tokens to enhance multilingual performance, Qwen 2.5 employs Byte-level BPE, utilising a 151,643-token vocabulary designed for multilingual processing, Gemma-2B has a SentencePiece tokeniser leveraging a 256,000-token vocabulary, making it highly effective for English-based tasks. Each model was fine-tuned on its respective downstream task following a systematic hyperparameter search to identify optimal configurations.
Prior to fine-tuning, prompt engineering was conducted to determine well-performing prompts tailored to each task, ensuring alignment with task-specific requirements and enhancing the models' ability to generate accurate and contextually relevant outputs. The hyperparameter search explored key factors, including weights for stability regularisation, mutual information (MI) regularisation, and the overall CARMA weight (Equation \ref{eq:carma_loss}), as well as the specific layers to which these losses were applied.

For training parameters, the following batch sizes were set in the IDM task: 16 for the Gemma-2B and GPT models, 32 for the Qwen-3B and Llama models, and 64 for the Qwen-0.5B model. For SC, the batch sizes were 16 for the GPT models, Gemma-2B and Llama-3B; 32 for Llama-1B and Qwen-3B; and 64 for Qwen-0.5B. For the number of training epochs, in the IDM, the Gemma and GPT models were trained for two epochs, while all other models were trained for three epochs, whereas all models were trained for two epochs, except Gemma-2B and LLama-1B, which were trained for three epochs for the SC task. The stopping layers for IDM and CARMA were configured as follows: GPT2-S at layer 3, GPT2-L at layer 8, Gemma-2B at layer 10, Llama-1B at layer 7, Llama-3B at layers 8 (stability) and 12 (MI), Qwen-0.5B at layer 5, and Qwen-3B at layer 10. The SC, the ending layers, 4 for GPT2-S, 12 for GPT2-L, 10 for Gemma-2B, 7, for LLama 1B, 8, for LLama 3B, 5 for Qwen-0.5B and 7 for Qwen-3B. For CARMA weight, optimal values varied by model size: 0.4 and 0.5 were most effective for larger models. We hypothesise that CARMA regularisation exhibits a weaker effect when lower weights are applied, particularly in larger architectures where stronger constraints are needed to stabilise compositional representations. In IDM, GPT2-L and Gemma performed best with a weight of 0.3, GPT2-S with 0.2, Llama-1B with 0.4, and Llama-3B with 0.5. Qwen models used 0.5 and 0.4 for the 0.5B and 3B variants, respectively. For SC Carma weight, it was 0.4 for Qwen-0.5B and GPT models, 0.5 for LLama-3B and Qwen-3B, and 0.3 for the rest. For the ending layer, it was 4 for GPT2-S, 12 for GPT2-L, 10 for Gemma-2B, 7 for LLama-1B, 8 for LLama-3B, 5 for Qwen-0.5B and 7 for Qwen-3B.

\begin{table}[ht]
\centering
\renewcommand{\arraystretch}{1.2} 
\setlength{\tabcolsep}{6pt} 
\resizebox{\columnwidth}{!}{%
\begin{tabular}{|l|c|c|c|c|c|c|}
\hline
\textbf{Model} & \textbf{Parameters} & \textbf{Layers} & \textbf{D\textsubscript{model}} & \textbf{Heads} & \textbf{Activation} & \textbf{MLP Dimension} \\
\hline
GPT-2 Small & 85M  & 12  & 768   & 12  & GELU & 3072  \\
GPT-2 Large  & 708M & 36  & 1280  & 20  & GELU & 5120  \\
Gemma-2B  & 2B   & 32  & 4096  & 16  & GELU & 8192  \\
LLaMA3.2 1B  & 1.1B & 16  & 2048  & 32  & SiLU & 8192  \\
LLaMA3.2 3B  & 3.2B & 28  & 3072  & 24  & SiLU & 8192  \\
Qwen2.5-0.5B  & 391M  & 24  & 896   & 14  & SiLU & 4864  \\
Qwen2.5-3B    & 3.0B  & 36  & 2048  & 16  & SiLU & 11008  \\
\hline
\end{tabular}%
}
\caption{Summary of model architectures. \textbf{Parameters}: total number of trainable parameters; \textbf{Layers}: total number of transformer layers; \textbf{D\textsubscript{model}}: size of word embeddings and hidden states; \textbf{Heads}: number of self-attention heads; \textbf{Activation}: activation function used in feedforward layers; \textbf{MLP Dimension}: dimensionality of the feedforward network.}
\label{tab:model_properties}
\end{table}

\subsection{Evaluation Metrics}\label{sec:evaluation_metrics}
This section details the evaluation metrics used in the study, including accuracy, synonym consistency, and performance stability.

\paragraph{Accuracy} is used as a primary measure of model performance and is defined as:

\begin{equation}
    \text{Accuracy} = \frac{TP + TN}{TP + TN + FP + FN},
\end{equation}

\noindent where $TP$ (true positives) and $TN$ (true negatives) denote correctly classified instances, while $FP$ (false positives) and $FN$ (false negatives) represent misclassified instances.

\paragraph{Synonym Consistency (ConsistSyn)} quantifies a model's ability to maintain correct predictions after synonym replacement. It is computed as:

\begin{equation}
    \text{ConsistSyn} = \frac{|\text{Correct After Replacement}|}{|\text{Correct Before Replacement}|} \times 100,
\end{equation}

\noindent where $\text{Correct After Replacement}$ refers to the number of correct predictions following synonym substitution, and $\text{Correct Before Replacement}$ denotes the number of correct predictions before substitution. The reported results are the averaged \text{ConsistSyn} across ($ N\geq 5$) runs. 

\paragraph{Coefficient of Variation (CV)} measures the stability of model performance across multiple runs, with lower values indicating greater consistency. It is defined as:

\begin{equation}
    \text{CV} = \frac{\sigma}{\mu},
\end{equation}

\noindent where $\sigma$ represents the standard deviation of model performance across runs, and $\mu$ denotes the mean performance.

\paragraph{Normalised Improvement (NI)} evaluates the relative gain in consistency introduced by a model over a baseline model. It is calculated as:

\begin{equation}
    \text{NI} = \frac{\text{ConsistSyn}_{\text{CARMA}} - \text{ConsistSyn}_{\text{baseline}}}{\text{ConsistSyn}_{\text{baseline}}} \times 100.
\end{equation}

\noindent This metric captures the percentage improvement in synonym consistency due to a model variant compared to the baseline model.

\subsection{Experimental setup}\label{sec:exp_setup}
Experiments were conducted using NVIDIA RTX A6000 and A100 GPUs. The method was developed in Python (v3.10.15) with Transformers (v4.44.2) \cite{wolf-etal-2020-transformers}, PyTorch (v2.4.1) \cite{paszke2019pytorch}, and Transformer-lens (v2.8.1) \cite{nanda2022transformerlens}. Preprocessing tasks, including tokenisation and tagging, used NLTK (v3.9.1) \cite{bird2009natural}, spaCy (v3.7.2) \cite{honnibal2020spacy}, and TextBlob (v0.18.0) \cite{loria2018textblob}, with Scikit-learn (v1.5.1) \cite{scikit-learn} for evaluation. Models use 500 warm-up steps and a 0.006 learning rate.

\section{Comprehensive Explanation of Evaluation Interventions}
\subsection{Constituent-Aware Pooling (CAP) Formalisation}\label{sec:cap_explanation}
Constituent-Aware Pooling (CAP) Formalisation is a method proposed in \cite{aljaafari2024interpreting} to systematically assess compositional generalisation via aggregating token-level activations into higher-level semantic representation. Below is a detailed explanation and formalisation of CAP. 
\paragraph{Overview.} CAP aggregates model activations at any chosen constituency level (e.g. tokens to words), enabling the analysis of compositional dependencies. The key steps involved are:
\begin{itemize}
    \item \textbf{Input Representations:} For a given input sequence \( X = [x_1, x_2, \ldots, x_n] \), the model produces inner states \( H = [h_1, h_2, \ldots, h_n] \) at a specific layer.  

    \item \textbf{Grouping Constituents:} Using syntactic parsers such as Benepar \cite{kitaev-etal-2019-multilingual, kitaev-klein-2018-constituency}, or by inversing the model tokeniser function, the sequence is segmented into constituents \( C = [c_1, c_2, \ldots, c_m] \), where each \( c_i \) represents a phrase or syntactic unit. For the experiments presented in the paper, tokens were grouped into words to form the smallest linguistic units.  

    \item \textbf{Pooling Operations:} For each constituent \( c_i \), the corresponding activations \( \{h_j | x_j \in c_i\} \) are aggregated into a single representation \( r_i \) using a pooling function:
    \[
    r_i = \alpha(\{h_j | x_j \in c_i\}) 
    \]
    CAP supports three pooling functions:
    \begin{itemize}
        \item \textbf{Maximum pooling:} Selects the highest activation values as: \[ \alpha(\{h_j | x_j \in c_i\}) = \max(\{h_j | x_j \in c_i\}), \quad\]
        \item \textbf{Mean pooling:} Computes the average of activation values as: \[\alpha(\{h_j | x_j \in c_i\}) = \frac{1}{|c_i|} \sum_{j \in c_i} \{h_j | x_j \in c_i\}, \quad\]
        \item \textbf{Sum pooling:} Accumulates activation values as: \[\alpha(\{h_j | x_j \in c_i\}) = \sum_{j \in c_i} \{h_j | x_j \in c_i\}.\]
    \end{itemize}

    \item \textbf{Updating Representations:} The pooled representations \( R = [r_1, r_2, \ldots, r_m] \) replace the original activations \( H \) for further processing.  
\end{itemize}

\begin{figure*}[h!]
    \centering
    \subfigure[GPT2-S]{
        \includegraphics[width=0.48\textwidth]{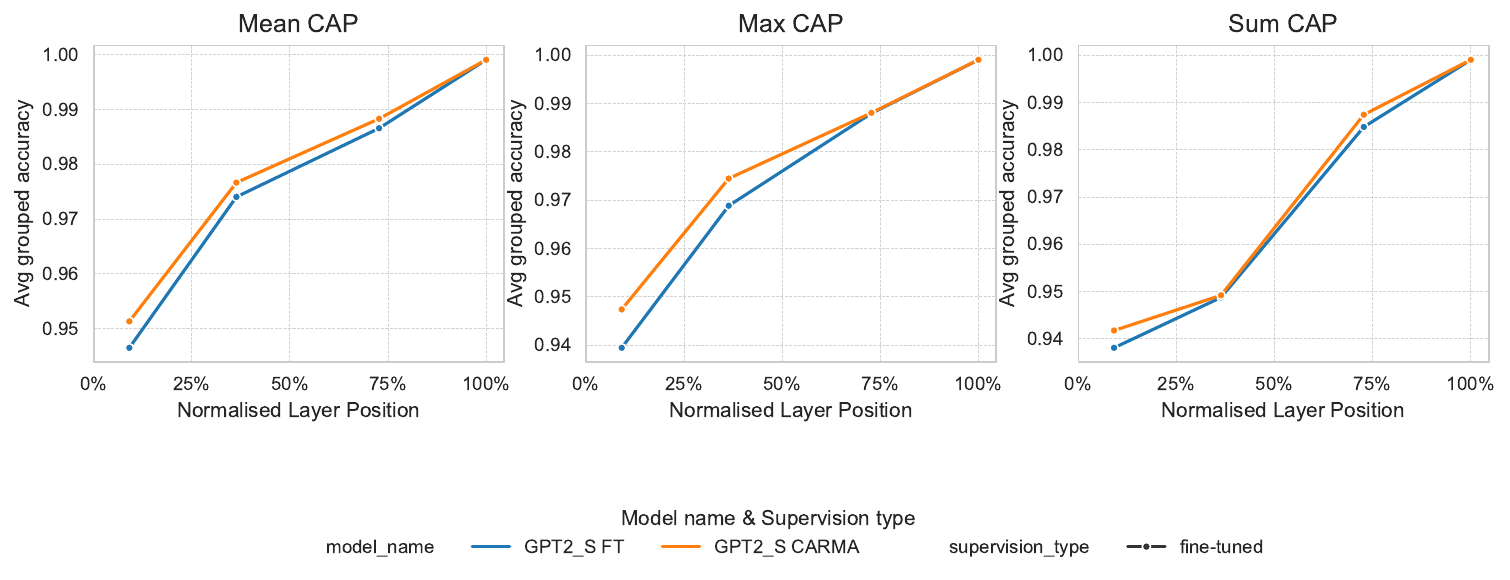}
    }
    \subfigure[GPT2-L]{
        \includegraphics[width=0.48\textwidth]{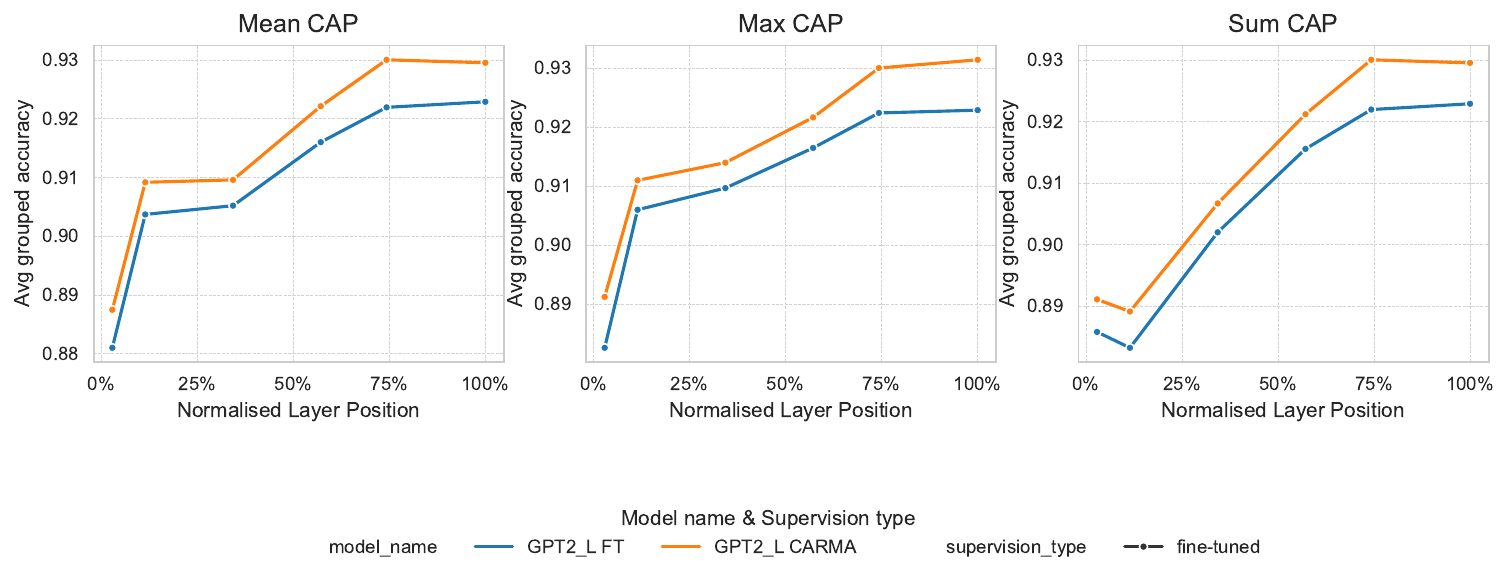}
    } \\
    \subfigure[Gemma-2B]{
        \includegraphics[width=0.48\textwidth]{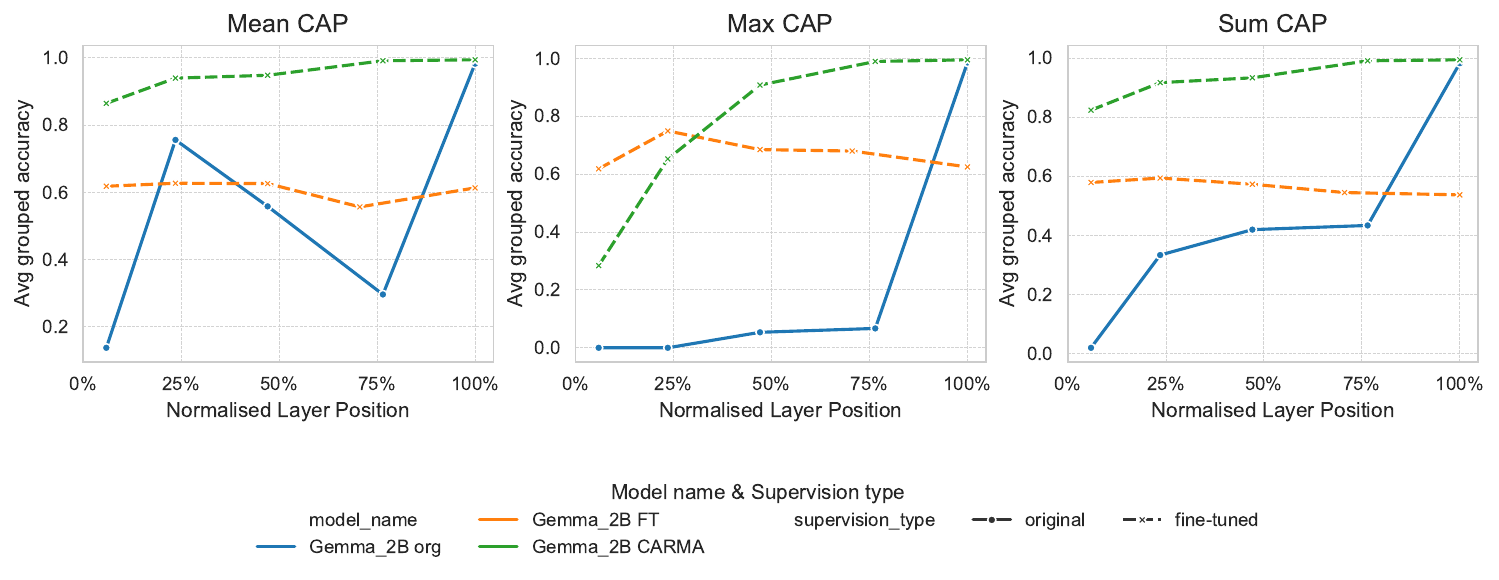}
    }
    \subfigure[Qwen-0.5B]{
        \includegraphics[width=0.48\textwidth]{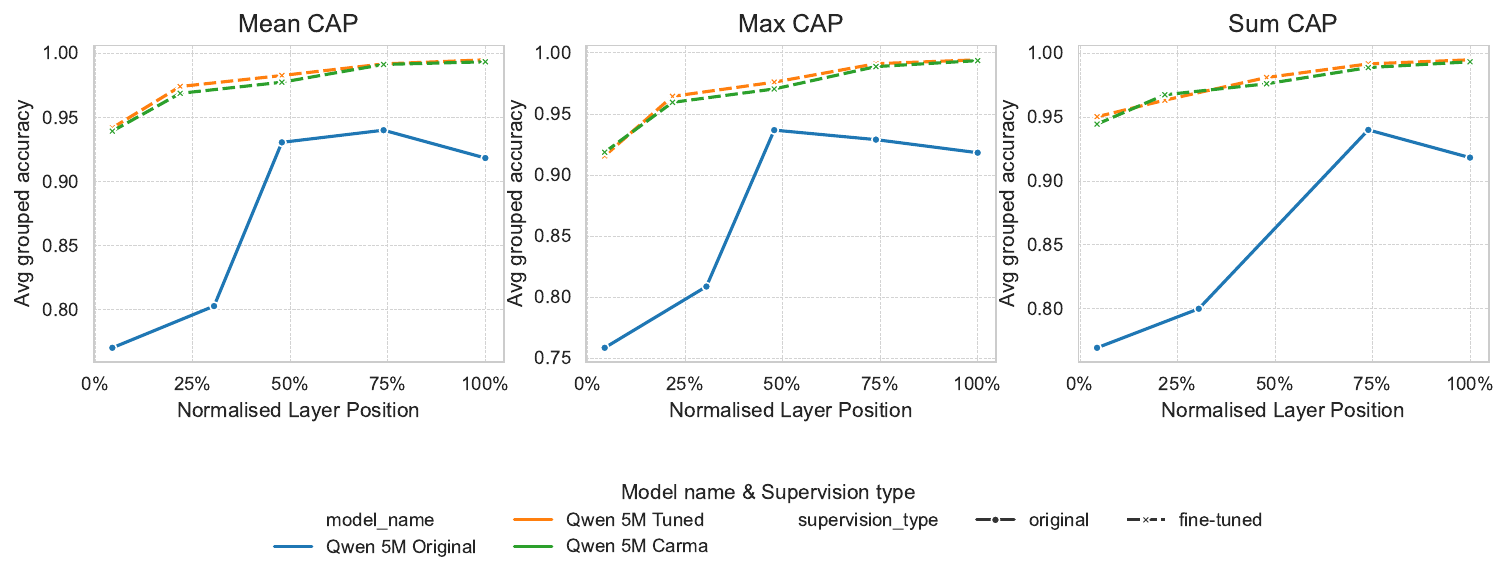}
    } \\
    \subfigure[Llama-1B]{
        \includegraphics[width=0.48\textwidth]{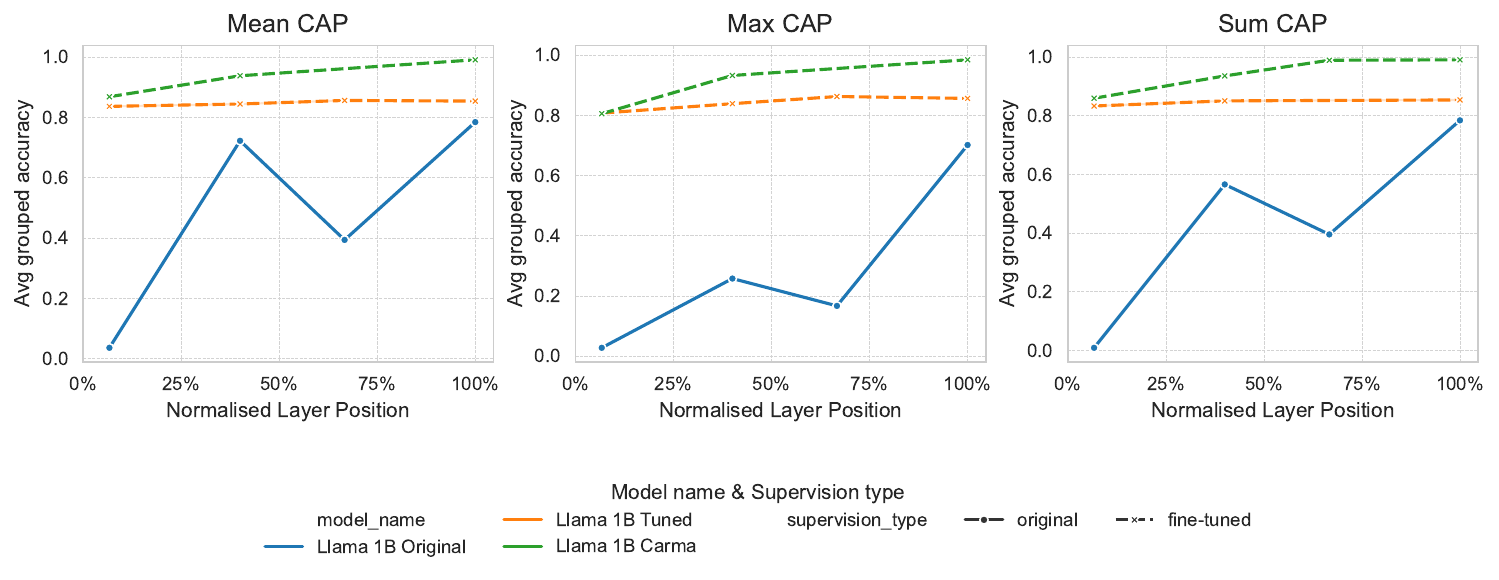}
    } 
    \subfigure[Llama-3B]{
        \includegraphics[width=0.48\textwidth]{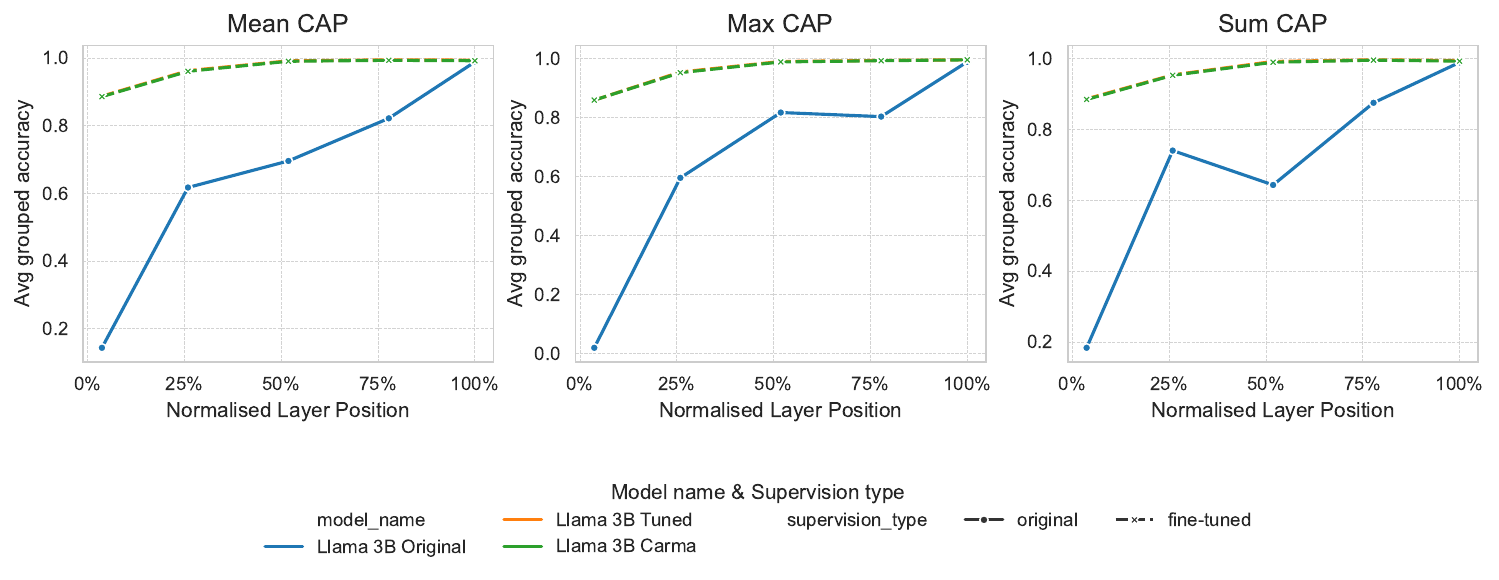}
    }
    \caption{SC Performance Across Models Under CAP}
    \label{fig:sst_models}
\end{figure*}

\paragraph{Evaluation.} The impact of CAP is evaluated by comparing task-specific performance metrics (e.g., accuracy, F1 score) of models before and after CAP is applied. This allows for a direct assessment of how CAP affects compositionality and task performance. This paper utilises the word-level CAP, pooling related token representation to their corresponding words. 
\begin{figure*}
    \centering
    \subfigure[IDM Task]{\label{fig:carma_results_idm_full}\includegraphics[width=.45\linewidth]{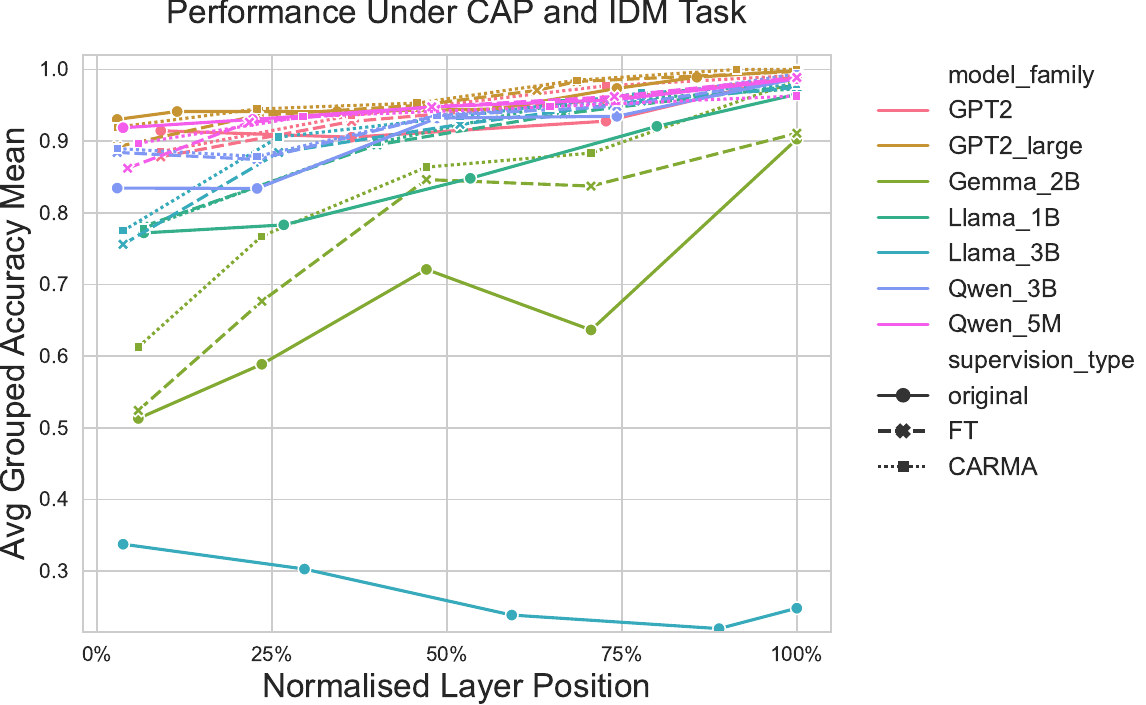}}
    \subfigure[SC Task]{\label{fig:carma_results_sst_full}\includegraphics[width=.45\linewidth]{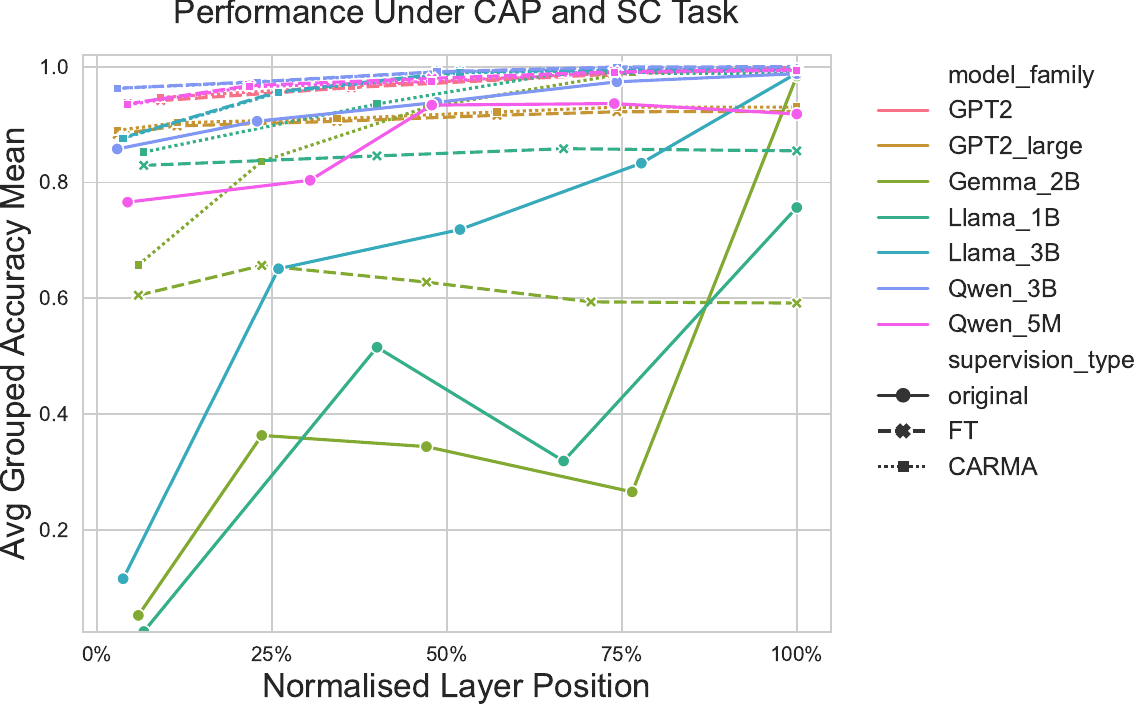}}
    \caption{Layer-wise performance under CAP interventions on the IDM (left) and SC (right) tasks. Results are averaged over three pooling strategies (Mean, Max, Sum) and reported for Original, Fine-Tuned (FT), and CARMA (FT + CARMA) models. Layer indices are normalised to support comparison across model sizes. CARMA improves robustness and systematicity in early-to-mid layers for both tasks, with diminishing differences in deeper layers.}
    \label{fig:carma_results_combined_full}
\end{figure*}
\begin{table}[h]
\centering
\small
\begin{tabular}{|l|c|c|c|c|c|}
\hline
\textbf{Model} & \textbf{Ver.} & \textbf{Task} & \textbf{Int.} & \textbf{CS} & \textbf{CV} \\
\hline
\multirow{6}{*}{GPT2-S} 
& CARMA & IDM & 25\% & 49.17 & 0.025 \\
& FT & IDM & 25\% & 50.89 & \textbf{0.017} \\
& Org & IDM & 25\% & \textbf{52.46} & 0.044 \\
\cline{2-6}
& CARMA & IDM & 40\% & 35.90 & \textbf{0.0542 }\\
& FT & IDM & 40\% & 37.16 & 0.0628 \\
& Org & IDM & 40\% & \textbf{37.20} & 0.1223 \\
\hline

\multirow{6}{*}{GPT2-L} 
& CARMA & IDM & 25\% & 56.31 & \textbf{0.0164} \\
& FT & IDM & 25\% & \textbf{56.95} & 0.0311 \\
& Org & IDM & 25\% & 51.10 & 0.1175 \\
\cline{2-6}
& CARMA & IDM & 40\% & 43.56 & {0.0485} \\
& FT & IDM & 40\% & \textbf{43.97} & \textbf{0.0459} \\
& Org & IDM & 40\% & 34.68  & 0.0895 \\
\hline

\multirow{6}{*}{Gemma-2B} 
& CARMA & IDM & 25\% & 56.70 & \textbf{0.023} \\
& FT & IDM & 25\% & \textbf{57.42} & 0.030 \\
& Org & IDM & 25\% & 49.47 & 0.031 \\
\cline{2-6}
& CARMA & IDM & 40\% & 0.4236 & \textbf{0.0174} \\
& FT & IDM & 40\% & \textbf{0.4498} & 0.0249 \\
& Org & IDM & 40\% & 0.3576 & 0.0480 \\
\hline

\multirow{6}{*}{Llama-1B} 
& CARMA & IDM & 25\% & \textbf{58.40 }& 0.0400 \\
& FT & IDM & 25\% & 57.86 &  \textbf{0.0385}\\
& Org & IDM & 25\% & 47.55 & 0.0503 \\
\cline{2-6}
& CARMA & IDM & 40\% & \textbf{47.07} & 0.0476 \\
& FT & IDM & 40\% & 46.75 & 0.0455 \\
& Org & IDM & 40\% & 33.49 & \textbf{0.0391} \\
\hline

\multirow{6}{*}{Qwen-0.5B} 
& CARMA & IDM & 25\% & \textbf{56.98} & 0.0286 \\
& FT & IDM & 25\% & 54.57 & \textbf{0.0191} \\
& Org & IDM & 25\% & 46.84 & 0.0684 \\
\cline{2-6}
& CARMA & IDM & 40\% & \textbf{40.55} & \textbf{0.0397} \\
& FT & IDM & 40\% & 39.69 & 0.0491 \\
& Org & IDM & 40\% & 32.98 & 0.0938 \\
\hline

\multirow{6}{*}{Qwen-3B} 
& CARMA & IDM & 25\% & \textbf{62.00} & \textbf{0.0225} \\
& FT & IDM & 25\% & 61.79 & 0.0279 \\
& Org & IDM & 25\% & 49.37& 0.0441 \\
\cline{2-6}
& CARMA & IDM & 40\% & 45.05 & \textbf{0.0400} \\
& FT & IDM & 40\% & \textbf{45.74} & 0.0551 \\
& Org & IDM & 40\% & 31.95 & 0.0688 \\
\hline

\multirow{6}{*}{Llama-3B} 
& CARMA & IDM & 25\% & \textbf{62.86} & \textbf{0.015} \\
& FT & IDM & 25\% & 62.22 & 0.029 \\
& Org & IDM & 25\% & 52.47 & 0.035 \\
\cline{2-6}
& CARMA & IDM & 40\% &\textbf{ 49.05 }& 0.0297 \\
& FT & IDM & 40\% & 48.31& \textbf{0.0191} \\
& Org & IDM & 40\% & 36.95 & 0.0458 \\

\hline
\end{tabular}
\caption{Model performance (25\% and 40\% synonym intervention) on the IDM task. \textbf{Ver.}: Version; \textbf{Int.}: Intervention rate; \textbf{CS}: ConsistSyn (\%); \textbf{CV}: Coefficient of Variation. \textbf{Best values in bold.}}
\label{tab:synonym_results_appendix_IDM}
\end{table}

\subsection{Synonym Replacement}\label{sec:syn_replacement}
A multi-step approach was adopted to ensure reliable synonym replacements. First, preprocessing was applied to filter out words that were unlikely to produce meaningful replacements. Specifically, words belonging to NLTK's predefined stopwords list or shorter than two characters were excluded from consideration. The remaining words were tagged with their part-of-speech (POS) using spaCy's \cite{honnibal2020spacy} POS tagger. Additionally, the sentiment of each word was determined using TextBlob \cite{loria2018textblob} to ensure that replacements preserved the semantic tone of the original text.
Next, a synonym vocabulary was constructed using words extracted from spaCy's $en\_core\_web\_md$ language model. This vocabulary was filtered to include only alphabetic common words with high probability scores (greater than -15 in our case), as determined by spaCy's word frequency data, while stopwords and rare terms were excluded. This step ensured that the vocabulary consisted of meaningful and contextually appropriate words for replacement.
For each target word, a list of synonym candidates was generated by iterating over the constructed vocabulary. The top $n$ candidates were selected based on their semantic similarity to the original word, measured using spaCy's word vectors. Synonyms with high similarity scores and alignment in POS were prioritised to maintain grammatical and contextual coherence in the text.
\begin{table}[]
\centering
\small
\begin{tabular}{|l|c|c|c|c|c|}
\hline
\textbf{Model} & \textbf{Ver.} & \textbf{Task} & \textbf{Int.} & \textbf{CS} & \textbf{CV} \\
\hline
\multirow{5}{*}{GPT2-S} 
& CARMA & SC & 25\% & 89.03 & 0.8903 \\
& FT & SC & 25\% & \textbf{89.54} & \textbf{0.8954} \\
\cline{2-6}
& CARMA & SC & 40\% &84.95 & \textbf{0.0095} \\
& FT & SC & 40\% & \textbf{85.07} & 0.0098 \\
\hline

\multirow{5}{*}{GPT2-L} 
& CARMA & SC & 25\% & \textbf{88.58} & \textbf{0.0065} \\
& FT & SC & 25\% & 88.04 & 0.0082 \\
\cline{2-6}
& CARMA & SC & 40\% & \textbf{84.61} & \textbf{0.0072} \\
& FT & SC & 40\% & 84.04 &0.0073  \\
\hline

\multirow{6}{*}{Gemma-2B} 

& CARMA & SC & 25\% & \textbf{84.81} & \textbf{0.0069} \\
& FT & SC & 25\% & 81.67 & 0.0088 \\
& Org & SC & 25\% & 68.14 & 0.0076 \\
\cline{2-6}
& CARMA & SC & 40\% & \textbf{81.48}  & 0.0102 \\
& FT & SC & 40\% & 74.29 & \textbf{0.0073} \\
& Org & SC & 40\% & 76.06 & 0.0136 \\
\hline

\multirow{6}{*}{Llama-1B} 
& CARMA & SC & 25\% & 74.03 & 0.0069 \\
& FT & SC & 25\% & \textbf{75.69} & \textbf{0.0044} \\
& Org & SC & 25\% & 2.65 & 0.1239 \\
\cline{2-6}
& CARMA & SC & 40\% & 71.43 & \textbf{0.0065} \\
& FT & SC & 40\% & \textbf{74.31} & 0.0102 \\
& Org & SC & 40\% & 1.73 & 0.2245 \\
\hline

\multirow{6}{*}{Qwen-0.5B} 
& CARMA & SC & 25\% & 89.66 & \textbf{0.0037} \\
& FT & SC & 25\% & \textbf{89.83} & 0.0085 \\
& Org & SC & 25\% & 59.12 & 0.0691 \\
\cline{2-6}
& CARMA & SC & 40\% & 86.03 & 0.0084 \\
& FT & SC & 40\% & \textbf{86.31} & \textbf{0.0046} \\
& Org & SC & 40\% & 55.27 & 0.0429 \\
\hline

\multirow{6}{*}{Qwen-3B} 
& CARMA & SC & 25\% & 93.65 & 0.0061 \\
& FT & SC & 25\% & \textbf{93.85} & \textbf{0.0039} \\
& Org & SC & 25\% & 67.63 & 0.0227 \\
\cline{2-6}
& CARMA & SC & 40\% & \textbf{91.26} & \textbf{0.0050} \\
& FT & SC & 40\% & \textbf{91.26 }& \textbf{0.0050} \\
& Org & SC & 40\% & 64.05 & 0.0159 \\
\hline

\multirow{6}{*}{Llama-3B} 
& CARMA & SC & 25\% & 84.83 & \textbf{0.0056} \\
& FT & SC & 25\% & \textbf{85.85} & 0.0065 \\
& Org & SC & 25\% & 35.21 &0.0136 \\
\cline{2-6}
& CARMA & SC & 40\% & 82.89 & \textbf{0.0016} \\
& FT & SC & 40\% & \textbf{83.55} & 0.0067 \\
& Org & SC & 40\% & 32.88 &0.0188 \\
\hline
\end{tabular}
\caption{Model performance (25\% and 40\% synonym intervention) on the SC task. \textbf{Ver.}: Version; \textbf{Int.}: Intervention rate; \textbf{CS}: ConsistSyn (\%); \textbf{CV}: Coefficient of Variation. \textbf{Best values in bold.}}
\label{tab:synonym_results_appendix_SC}
\end{table}
\section{InfoNCE for Mutual Information Estimation}\label{sec:infoNCE_MI}
Mutual information (MI) quantifies the shared information between two variables \(X\) and \(Y\). CARMA leverages MI maximisation to capture dependencies between tokens effectively, thereby enhancing compositional generalisation in LLMs. Specifically, CARMA uses MI, denoted as \(I(X; Y)\), to reinforce token-level interactions critical for compositionality. However, direct computation of MI is challenging in practice.

To address this challenge, a variant of InfoNCE is employed to estimate MI and approximate these dependencies efficiently. Given an anchor token hidden state \(h_i\), we construct a corresponding positive set \(\mathbf{H}\), which contains tokens hidden states semantically or syntactically related to \(h_i\). Additionally, we define \(\mathcal{N}\) as the set of negative examples consisting of unrelated tokens hidden states.

The InfoNCE objective provides a practical lower bound on \(I(X; Y)\) \cite{oord2018representation}, as follows:
\begin{equation}
\small
    I(X; Y) \geq \mathbb{E} \left[ \log \frac{\sum_{h_j \in \mathbf{H}} f(h_i, h_j)}{\sum_{h_j \in \mathbf{H}} f(h_i, h_j) + \sum_{h_k \in \mathcal{N}} f(h_i, h_k)} \right],
\end{equation}
where \(f(h_i, h_j) = \exp(\text{sim}(h_i, h_j) / \tau)\) is a scaled similarity function, and \(\tau\) is a temperature parameter. This adaptation of InfoNCE introduces token-specific interactions within the layer-wise structure of LLMs, ensuring that dependencies are captured across layers. By maximising mutual information, CARMA aligns the optimisation direction to enhance compositional structures.

To extend this approach across layers, the final CARMA MI loss is computed as:
\begin{equation}
\small
\begin{aligned}
    \mathcal{L}_{\text{MI}} = - \frac{1}{N} \sum_{i=1}^N \Bigg( 
    &\log \sum_{\substack{h_j \in \mathbf{H} \\ j \neq i}} \exp\left(\frac{\text{sim}(h_i, h_j)}{\tau}\right) \\
    &- \log \Bigg( \sum_{\substack{h_j \in \mathbf{H} \\ j \neq i}} \exp\left(\frac{\text{sim}(h_i, h_j)}{\tau}\right) \\
    &\quad + \sum_{h_k \in \mathcal{N}} \exp\left(\frac{\text{sim}(h_i, h_k)}{\tau}\right) \Bigg) \Bigg),
\end{aligned}
\end{equation}
where \(h_i\) is the anchor token, \(h_j \in \mathbf{H}\) are positive examples related to \(h_i\), \(h_k \in \mathcal{N}\) are negative examples, \(N\) is the number of anchors, and \(\text{sim}(h_i, h_j)\) is a similarity function. The negative sign ensures that MI is maximised during optimisation. Without this negative sign, the objective would incorrectly minimise MI, thereby hindering CG enhancement.

\section{Extended results}\label{sec:additiona_results}
Figures~\ref{fig:idm_models}, \ref{fig:sst_models} and Tables~\ref{tab:synonym_results_appendix_IDM} and~\ref{tab:synonym_results_appendix_SC} provide additional results for models' performance comparison under CAP and synonym interventions. Figure \ref{fig:carma_results_combined_full} shows an overall models performance under cap for all models. CARMA models show a clear advantage over all models and tasks. However, the gain is clearer in the IDM case, where more intricate features and compositionality generalisation are required. It is also observed that the performance of the FT and CARMA models demonstrates similar curves or trends. Given this observation, we argue that CARMA's improvements stem from its learning objectives, which align closely with cross-entropy loss while explicitly addressing intermediate representation stability. The observed improvements are moderate in some cases, particularly for SC tasks. This behaviour is expected due to the limited size of the fine-tuning datasets compared to the original pretraining data used for these models. Nevertheless, larger models, such as Llama-3B and Gemma-2B, exhibit more substantial improvements with CARMA, demonstrating its scalability with model capacity.


\subsection{Training Runtime and Overhead}
\label{app:carma_overhead}

We report wall-clock training times (in minutes) for each model under standard fine-tuning (FT) and with \textsc{CARMA}. Overhead is computed as the relative increase in runtime caused by the additional mutual information and stability losses. All experiments for models were conducted on a single GPU under identical batch size, optimiser settings, and hardware configuration. 
\begin{table}[h]
\centering
\small
\begin{tabular}{lccc}
\toprule
\textbf{Model} & \textbf{FT (min)} & \textbf{CARMA (min)} & \textbf{Overhead ratio} \\
\midrule
GPT2 Small       &   1.9 & 3.48  & $\times 1.8$ \\
GPT2 Large        &   6.7 &  8.9 & $\times 1.3$ \\
Llama 3.2–1B       &   2.78  &  6.96 & $\times 2.5$ \\
Llama 3.2–3B       &   7.55  &  20.2 & $\times 2.2$ \\
Qwen 2.5–0.5B      &  2.10 &  4.75  & $\times 2.3$ \\
Qwen 2.5–3B        &  2.98 &  5.01 & $\times 1.7$  \\
\bottomrule
\end{tabular}
\caption{Wall-clock IDM training time and overhead introduced by CARMA. All runs use a single GPU under identical batch, optimiser, and hardware settings.}
\end{table}

\begin{table}[h]
\centering
\small
\begin{tabular}{lccc}
\toprule
\textbf{Model} & \textbf{FT (min)} & \textbf{CARMA (min)} & \textbf{Overhead ratio} \\
\midrule
GPT-2 Small       &   1.12 & 8.51  & $\times7.6$ \\
GPT-2 Large        &  6.02  & 13.18 & $\times2.19$ \\
LLaMA 3.2–1B       &   2.50 & 9.98  & $\times4.9$ \\
LLaMA 3.2–3B       &  6.58 &  14.06  & $\times2.13$ \\
Qwen 2.5–3B        &  6.01 & 16.86  & $\times 2.8$ \\
\bottomrule
\end{tabular}
\caption{Wall-clock SC training time and overhead introduced by CARMA. All runs use a single GPU under identical batch, optimiser, and hardware settings.}
\end{table}

\textsc{CARMA} introduces non-trivial training-time overhead due to auxiliary objectives, particularly for smaller models or longer sequences. However, inference costs remain unchanged, and no architectural modifications are required. We observe moderate to high training slowdowns (e.g., $\times$2.2–$\times$2.9 for LLaMA-3B), with variance across models due to token length and loss computation. Optimising runtime for mutual information and stability estimation is an important direction for future efficiency improvements.

\end{document}